# Bridging the Skill Gap in Clinical CBCT Interpretation with CBCTRepD

Qinxin Wu[1], Fucheng Niu[2], Hengchuan Zhu[2], Yifan Sun[1], Ye Shen[1], Xu Li[1], Han Wu[1], Leqi Liu[1], Zhiwen Pan[1], Zuozhu Liu[2,✉], Fudong Zhu[1,✉], and Bin Feng[1,✉]

[1] Stomatology Hospital, School of Stomatology, Zhejiang University School of Medicine, Zhejiang Provincial Clinical Research Center for Oral Diseases, Zhejiang Key Laboratory of Oral Biomedical, Hangzhou, China.

[2] College of Computer Science and Technology, Zhejiang University-University of Illinois Urbana-Champaign Institute, Zhejiang University, Hangzhou 310027, Zhejiang, China.

✉Corresponding author

## Abstract

Generative AI has advanced rapidly in medical report generation; however, its application to oral and maxillofacial CBCT reporting remains limited, largely because of the scarcity of high-quality paired CBCT–report data and the intrinsic complexity of volumetric CBCT interpretation. To address it, we introduce CBCTRepD, a bilingual oral and maxillofacial CBCT report-generation system designed for integration into routine radiologist–AI co-authoring workflows. we curated a large-scale, high-quality paired CBCT–report dataset comprising approximately 7,408 studies, covering 55 oral disease entities across diverse acquisition settings, and used it to develop the system. We further established a clinically grounded, multi-level evaluation framework that assesses both direct AI-generated drafts and radiologist-edited col laboration reports using automatic metrics together with radiologist- and clinician-centred evaluation. Using this framework, we show that CBCTRepD achieves superior report-generation performance and produces drafts with writing quality and standardization comparable to those of intermediate radiologists. More importantly, in radiologist–AI collaboration, CBCTRepD provides consistent and clinically meaningful benefits across experience levels: it helps novice radiologists improve toward intermediate-level reporting, enables intermediate radiologists to approach senior-level performance, and even assists senior radiologists by reducing omission-related errors, including clinically important missed lesions. By improving report structure, reducing omissions, and promoting attention to co-existing lesions across anatomical regions, CBCTRepD shows strong and reliable potential as a practical assistant for real-world CBCT reporting across multi-level care settings.

## Introduction

Oral and maxillofacial disorders encompass diverse, high-burden conditions managed across oral and maxillofacial surgery, endodontics, orthodontics and periodontal–implant–prosthodontic care.[1] Cone-beam computed tomography (CBCT) is central to diagnosis and treatment planning because it provides high-resolution three-dimensional depiction of dentoalveolar and craniofacial anatomy, often with multiple co-existing findings across the jaws, temporomandibular joint and paranasal sinuses.[2, 3] Translating these volumes into a structured radiology report remains a bottleneck. Report writing is time-consuming, strongly experience-dependent and demands precise spatial reasoning over densely packed hard–soft tissue interfaces, vigilance for subtle lesions and standardized communication of clinically actionable findings.[4, 5] This challenge is amplified by the limited availability of dedicated oral and maxillofacial radiologists, particularly in novice-care settings, where report quality can vary and safety-relevant omissions are a persistent concern.

Generative AI has rapidly advanced medical report generation through large language models and vision–language models, yet clinical deployment remains constrained by known failure modes, including hallucinated content, terminology bias, limited access to high-quality paired data and performance drift under distribution shifts.[6-12] These limitations are especially salient for dental CBCT. The task requires volumetric reasoning, accurate tooth- and region-level localization, and consistent coverage of multi-lesion presentations under heterogeneous clinical indications[13-17]. Despite the ubiquity of CBCT, dedicated solutions for bilingual CBCT report drafting that can be integrated into routine practice remain scarce[18-20].

This study is motivated by several persistent challenges in CBCT report generation.[21] The systematic development of CBCT-specific medical report generation (MRG) remains limited, hindered by the scarcity of large-scale, high-quality datasets. Existing resources are insufficient to capture the volumetric complexity, multi-region coverage, and long-tail distributions of oral and maxillofacial diseases, which constrains model performance in real-world clinical scenarios. [22, 23] Moreover, most prior MRG studies have not been evaluated within actual clinical workflows, leaving their practical utility and integration potential largely unexplored. In addition, conventional evaluation approaches remain narrow and imprecise, often relying solely on string-based metrics that fail to assess diagnostic accuracy, clinical relevance, or the impact on reporting workflows and safety-critical outcomes. Taken together, these limitations underscore the urgent need for a comprehensive, clinically grounded framework for CBCT report generation.

Building on these challenges, we present CBCTRepD, a vision–language model specifically designed for bilingual CBCT report generation, and, for the first time, deployed it within real-world radiologist–AI collaborative workflows. In this setting, radiologists review and refine AI-generated drafts to produce final reports, allowing direct evaluation of clinical utility while preserving the natural sequence of diagnosis, imaging, and reporting. To enable CBCTRepD, we curated the first large-scale, high-quality CBCT–report dataset, comprising 7,408 paired studies spanning 55 complex oral disease entities, covering multiple tissues and anatomical sites, including bone, teeth, and sinuses. Leveraging this dataset, CBCTRepD consistently outperforms prior general-purpose and medical-specific vision–language models, particularly in identifying coexisting lesions and generating accurate, clinically coherent reports. In routine workflow simulations, the model produces intermediate-comparable drafts that match or surpass manual reports from less-experienced radiologists and approach the quality of senior specialists. Critically, our study demonstrates that based on CBCTRepD, human–AI collaboration enhances reporting capability across all experience tiers: Novice radiologists achieve substantial improvements, reaching intermediate-level proficiency; Intermediate radiologists approach senior-level performance; and even Senior radiologists benefit from measurable gains.

The core innovations of our work include: (1) the first AI system integrating CBCT report generation directly into clinical practice routine; (2) a realistic, end-to-end clinical evaluation framework that comprehensively assesses report quality, safety, and usability across multiple radiologist tiers; (3) a pioneering, large-scale dataset of complex multi-disease CBCT scans, rich in anatomical diversity and coexisting lesions; (4) robust performance gains over existing general and medical vision–language models, with particular advantage in co-disease detection; (5) demonstrated benefits of CBCTRepD-assisted collaboration reporting in reducing omissions, standardizing structure, and improving clinical usability across radiologists of varying experience. Collectively, CBCTRepD establishes a principled framework for clinical deployment of AI-assisted reporting, closing skill gaps across all radiologist tiers while enhancing report completeness and clinical utility. It provides a foundation for future CBCT report generation and can be applied in real-world clinical workflows in a repeatable and scalable manner.

## Results

### Overview of CBCTRepD and clinical workflow

We constructed a high-quality, multi-disease CBCT-report paired dataset as the training foundation (Fig. 1a, Fig. 2; Table S1). Based on this dataset, we developed CBCTRepD, a bilingual AI system designed to generate dental CBCT radiology reports. The system integrates CBCT imaging with clinicians' preliminary assessments and employs a vision-language architecture that fuses visual and textual annotations to produce reports (Fig. 1c), encompassing key clinical content such as imaging findings and diagnostic impressions. The generated reports are subsequently refined by radiologists, forming human-AI collaborative reports to support patient consultation and clinical decision-making (Fig. 1b; Fig. S1). Using a realistic novel framework, we conducted comprehensive evaluations of CBCTRepD in fully simulated clinical environments, assessing both performance and collaborative effectiveness from upstream and downstream tasks as well as objective metrics (Fig. 1d; Fig. S1).

### Dataset overview

Dental CBCT volumes depict dense oral–maxillofacial anatomy and encompass heterogeneous disease presentations. To capture this complexity, we curated a large-scale paired dataset linking CBCT volumes, clinical diagnoses, and radiology reports, covering 55 oral disease entities (Table S8). In total, 7,108 CBCT–report pairs were included for model development (training cohort), and an independent 300 pair cohort was constructed for formal evaluation (test cohort) (Fig. 2).

Diagnostic entities were extracted from the Impressions section, revealing a long-tail distribution of impression entities across both cohorts (Fig. 2a–b). Cases were further organized by ordering departments and stratified by field-of-view to support subgroup analyses; detailed cohort characteristics and distributions are provided in Fig. 2 and Table S1.

### Comparison with Medical and General VLMs

We next test our method on both Chinese (ZH) and English (EN) report-generation tasks against open-source general-purpose VLMs and prior 3D medical VLMs, and our method achieved the best overall performance among all compared baselines (Fig. 3a–b; Supplementary Tables S2–S3). In ZH, our model achieved BLEU-4 = 0.311 together with ROUGE-L = 0.497, METEOR = 0.528 and BERTScore = 0.825, whereas the strongest 3D medical baseline (Med3DVLM-7B-FT) reached BLEU-4 = 0.220, ROUGE-L = 0.406, METEOR = 0.423 and BERTScore = 0.793, indicating consistent improvements across n-gram overlap, sequence-level similarity and semantic alignment. General-purpose VLM baselines showed markedly lower text similarity, suggesting limited transferability to CBCT reporting without domain adaptation (Table S2). A similar pattern was observed in EN. Our model obtained BLEU-4 = 0.126, ROUGE-L = 0.371, METEOR = 0.341 and BERTScore = 0.892, while Med3DVLM-7B-FT achieved BLEU-4 = 0.077, ROUGE-L = 0.302, METEOR = 0.275 and BERTScore = 0.881. Qualitative examples further illustrate that our model generates bilingual Findings and Impressions that more closely match expert reports than representative baselines and remains robust on long-tail disease cases (Fig. S2–S3). Collectively, these objective results demonstrate robust bilingual report generation quality and establish a strong basis for subsequent clinical utility and human–AI collaboration evaluations (Fig. 3; Table S2–S3).

**Comparison with Manual Reports from Three Tiers of Radiologists**

We next evaluated CBCTRepD reports alongside fully manual reports produced by radiologists stratified into three experience-level cohorts (Novice, Intermediate, and Senior) using complementary automatic metrics, expert-centred radiologist and clinician preferences, and dimension-wise quality scoring (Fig. 4; Tables S3–S5). CBCTRepD showed strong agreement with expert-style reporting at the linguistic level (Fig. 4a), achieving BLEU-4 = 0.30 and ROUGE-L = 0.47, exceeding the Novice cohort (0.07; 0.28) and approaching the Intermediate cohort (0.23; 0.41), while the Senior cohort remained closest to expert references (0.36; 0.54).

Quality score evaluation highlighted clinically relevant strengths (Fig. 4c). Radiologist scoring showed that CBCTRepD produced substantially more coherent reports than Novice reports (2.10 versus 3.35, lower is better), while maintaining clinical-use scores that were comparable to Intermediate reports (2.91 versus 2.42) and closer to the Senior profile than the Novice profile. Although Senior reports achieved the best scores across dimensions, CBCTRepD consistently tracked the Intermediate cohort, supporting its role as a intermediate-level drafting assistant that standardizes structure and reduces documentation burden. Clinician scoring exhibited a similar pattern with a compressed dynamic range, with CBCTRepD remaining competitive on coherence (2.40) and clinical use (2.66), and showing improved alignment over low-seniority manual reports (Fig. 4c).

To capture safety-relevant failure modes, we additionally summarized omission/incorrection behaviours (Fig. 4d; Tables S4–S5). CBCTRepD showed a lower omission burden than Novice reports for both Findings and Impression, and the omitted items were less frequently judged clinically significant (Findings 19% and Impressions 42%, compared with Novice 69% and 61%). Incorrections followed the same overall trend (Tables S4–S5), remaining above senior-level performance but supporting the view that CBCTRepD drafts are safer and more reliable than low-seniority manual reports, while still benefiting from expert review in high-stakes cases.

**The Role of Human-AI Collaboration in Enhancing the Quality of CBCT Reports**

We next evaluated whether AI-assisted reporting (Co-Novice, Co-Intermediate and Co-Senior) improves report quality relative to the corresponding manual baselines (Novice, Intermediate and Senior) across automatic metrics, preference ranking, quality scores and omission-related safety outcomes (Fig. 5; Tables S6–S8). The bilingual generation, manual, and collaboration reports comparison is shown in Fig. 6 (EN) and Fig.S6 (ZH).

We next examined how human–AI collaboration reports (radiologists editing CBCTRepD drafts) reshaped manual reporting quality across three experience-level cohorts (Novice, Intermediate and Senior) (Fig. 5; Tables S6–S8). Collaboration yielded consistent linguistic gains, with the largest uplift in the Novice cohort (BLEU-4: +129%, ROUGE-L: +29%), followed by moderate improvements in the Intermediate cohort (+16–35% across BLEU; ROUGE-L +20%) and smaller but measurable gains in the Senior cohort (+11–17% across BLEU; ROUGE-L +11%) (Fig. 5a). These objective improvements were mirrored by a clear preference shift: collaborative reports were less frequently assigned the lowest ranks and more frequently moved into higher-preference tiers across both radiologist and clinician raters (Fig. 5b; Table S6).

Quality-score evaluation further highlighted the practical value of AI assistance (Fig. 5c). Relative to each cohort's baseline manual reports, collaboration produced broad score reductions (improvements) across key dimensions—most prominently for coherence and clinical use in the Novice cohort (radiologists: Δcoherence −1.05, Δclinical use −0.45; clinicians: Δcoherence −0.34, Δclinical use −0.43), with smaller yet consistent improvements for Intermediate and Senior cohorts across factual consistency and medical safety (Fig. 5c).

Finally, collaboration improved safety-relevant completeness, with a robust reduction in omissions (Fig. 5d; Table S7), particularly for the Novice cohort (Δomission count: Findings −0.73, Impressions −0.28; Δclinically significant omissions: Findings −15.0%, Impressions −10.0%), and additional decreases in the Intermediate and Senior cohorts. In contrast, incorrections showed a mixed pattern across subspecialties (Table S8), decreasing in some settings while modestly increasing in others, indicating that collaboration primarily strengthens structure, completeness and usability, while residual factual verification remains an important role for expert review.

**Correlation between objective metrics and expert assessments**

We assessed how well objective NLG metrics reflect human judgement by computing Spearman correlations between BLEU/ROUGE-L and preference/quality evaluations from radiologists and clinicians, together with radiologist-recorded omission and incorrection burdens, with subspecialty-stratified correlation patterns reported in Supplementary Figs. S4–S5 (Fig. 4–5; Fig. S4–S5). In the radiologist total analysis, BLEU/ROUGE-L showed a clear monotonic association with safety-relevant error burden, with stronger correlations for omissions ($\rho = -0.45$ to $-0.48$ across Findings/Impressions) than for incorrections ($\rho = -0.32$ to $-0.36$), and a moderate association with preference ranking ($\rho = -0.34$ to $-0.39$), consistent with the omission reductions and preference shifts observed after human–AI collaboration (Fig. 5b,d). In the clinician total analysis, NLG metrics tracked overall acceptability more strongly, with preference ranking showing higher correlations ($\rho = -0.53$ to $-0.56$), and they additionally aligned with dimension-wise quality scores (for example, factual consistency $\rho = -0.43$ to $-0.45$, coherence $\rho = -0.44$ to $-0.50$, medical safety $\rho = -0.45$ to $-0.48$, and clinical use $\rho = -0.48$ to $-0.49$). Together, these results indicate that BLEU/ROUGE-L serve as useful high-throughput proxies for report preference and completeness, while expert/clinical scoring remains essential to capture safety- and correctness-critical aspects beyond surface-form similarity (Fig. 4c; Fig. 5c).

## Discussion

VLMs have shown promise in medical imaging, particularly in 2D modalities. By contrast, CBCT report generation in oral and maxillofacial practice remains underexplored, despite the clinical need for high-quality reporting in a setting marked by dense anatomy, volumetric complexity and frequent coexistence of multiple abnormalities. Here we present CBCTRepD, a bilingual system that generates complete CBCT reports, including Findings and Impression, from high-resolution oral and maxillofacial CBCT together with the initial clinical diagnosis. Beyond benchmark evaluation, we assessed the system in a simulated workflow designed to approximate routine reporting practice. Within this framework, CBCTRepD approached the performance range of intermediate readers on the evaluated tasks and improved collaborative reporting by increasing completeness and structural Coherence, particularly among less-experienced radiologists. By combining assessments from specialty clinicians and expert OMRs, we further developed a clinically grounded framework for evaluating report quality, usability and potential safety concerns in both standalone and collaborative settings.

To support model development, we constructed a large-scale maxillofacial CBCT–report paired dataset with substantial anatomical diversity and complex coexisting diseases. Our dataset enabled training and evaluation under more representative conditions and may have contributed to the broader diagnostic coverage of CBCTRepD. This feature may partly explain the stronger performance of CBCTRepD relative to existing medical and general-purpose VLMs in multi-disease scenarios.

Improved generation performance alone, however, does not establish clinical usefulness. The more relevant questions are whether generated reports are clinically usable, and whether AI assistance improves reporting within physician workflows. To address this, we designed an end-to-end evaluation spanning radiologists at three experience levels and assessed both standalone generated reports and human–AI collaboration. Under this framework, CBCTRepD-assisted reporting reduced experience-related variation, with the largest gains observed among novice and intermediate OMRs. These improvements were most evident in coherence, completeness and the inclusion of findings across a broader set of anatomically relevant regions, suggesting that the system functions less as a substitute for expert interpretation than as a structured assistant supports more standardized drafting and more comprehensive review.

Our results also propose a dissociation between language-based metrics and clinically assessment. Although CBCTRepD achieved natural language metric performance approaching that of senior OMRs, this was not matched by expert clinical scoring. Templated phrasing may partly explain this gap by improving sentence form similarity without ensuring lesion-level correctness, localized precision. Consistent with this, clinician-rated quality correlated relatively weak with conventional language metrics. Progress in medical report generation should therefore be judged primarily by clinically meaningful criteria, rather than by textual similarity alone.

Several limitations warrant consideration. Tooth-position localisation and fine-grained anatomical attribution remain important failure modes, and rare diagnostic categories are still underrepresented. More broadly, external validity remains uncertain across institutions, scanners, acquisition protocols and reporting cultures. In addition, our workflow evaluation was retrospective and simulated rather than prospective, and therefore cannot fully capture the effects of real-world user behaviour. Although the system is bilingual, equivalent clinical assessment across languages was not exhaustively established.

Taken together, these findings support the feasibility of clinically oriented AI assistance for CBCT reporting while clarifying the work needed before routine deployment. More broadly, this work illustrates how progress in medical report generation may depend not only on model design, but also on clinically representative data and evaluation frameworks. Prospective multi-centre validation, stronger anatomical grounding and explicit safety monitoring will be necessary to determine whether such systems can be translated reliably into routine medical practice.

## Methods

### Data Acquisition

An overview of the data collection and preprocessing workflow is illustrated in Fig. 1a. We retrospectively collected 7,408 cone-beam computed tomography (CBCT) examinations from 7,303 patients at the Stomatology Hospital, Zhejiang University School of Medicine, along with corresponding radiology reports and initial clinical diagnoses. Each report consisted of two structured sections: Findings, detailing descriptive imaging observations, and Impression, summarizing diagnostic conclusions.

After curation, 7,108 CBCT–report pairs were designated for model development (training cohort), while an independent set of 300 pairs was reserved for evaluation (test cohort) (Fig. 2). Diagnostic entities were extracted from the Impressions section to generate an entity-frequency profile, showing a long-tail distribution: a few high-frequency entities accounted for most occurrences, while many low-frequency entities appeared sporadically (Fig. 2a–b). Individual reports could contain multiple entities, total entity counts exceeded the number of CBCT studies.

In the training cohort, the most frequent impression entities included impacted tooth (4,896 occurrences), apical periodontitis (4,766), malocclusion (3,498), and dental caries (2,248), followed by post-treatment included post–root canal treatment (1,938), partial edentulism (1,928) (Fig. 2a). The test cohort preserved a similar clinical spectrum, with differences in relative entity frequency, including malocclusion (294), dental caries (248), sinusitis (238), altered TMJ space (210), and impacted tooth (194) being most frequent, reflecting the targeted sampling strategy.

### Model Architecture

**CBCTRepD** is a CBCT medical report generation vision–language model (VLM) designed to generate complete radiology reports by jointly leveraging CBCT imaging and clinical context. The model consists of four core components: (1) a rotary position-adaptive visual encoder, (2) a text tokenizer, (3) a multimodal projector, and (4) an LLM decoder.

**Rotary Position-Adaptive Visual Encoder.** The visual processing pipeline begins with a rotary position-adaptive visual encoder tailored for CBCT volumes. Given a CBCT scan, we represent the volume as an ordered sequence of 2D image planes extracted from the reconstructed volume. Each plane is partitioned into a grid of non-overlapping fixed-size image patches, which are linearly projected into patch tokens and processed by a Vision Transformer (ViT). The resulting per-plane token sequences are concatenated across planes to form the visual token stream for multimodal conditioning.

To encode spatial position in a resolution-agnostic manner, we replace learned absolute positional embeddings with 2D rotary positional embeddings (2D RoPE). For a patch located at grid coordinates $(m, n)$, we apply 1D RoPE independently along the height and width dimensions to the query and key vectors used in self-attention. Let $q \in \mathbb{R}^d$ denote a query (or key) vector with even dimension d. We partition the d features into $d/2$ pairs, where dimensions $2i-1$ and $2i$ form the $i$-th pair for $i \in \{1, \ldots, d/2\}$. For each spatial coordinate $p \in \{m, n\}$, we apply a rotation to each pair:

$$\begin{pmatrix} q'_{2i-1} \\ q'_{2i} \end{pmatrix} = \begin{pmatrix} \cos(p\theta_i) & -\sin(p\theta_i) \\ \sin(p\theta_i) & \cos(p\theta_i) \end{pmatrix} \begin{pmatrix} q_{2i-1} \\ q_{2i} \end{pmatrix},$$

where the frequencies are $\theta_i = 10000^{-2i/\mathrm{d}}$. We allocate half of the rotated dimensions to encode the height coordinate m and the remaining half to encode the width coordinate n. This formulation injects relative spatial information directly into the attention mechanism without requiring learned positional embeddings, enabling the encoder to accommodate CBCT planes with varying resolutions and fields of view.

**Text Tokenizer** For textual input, we employ the tokenizer native to the LLM backbone, which is a Byte-Pair Encoding (BPE) tokenizer with a vocabulary size of 152,064 tokens.[24]

**Multimodal Projector** To bridge the visual and text domains, a multimodal projector $(\mathrm{g}(\cdot))$ aligns the output of the vision encoder with the LLM's embedding space with a two-layer Multilayer Perceptrons (MLPs). It takes the final sequence of visual patch embeddings from the vision encoder $f_v(\mathbf{v}) \in \mathbb{R}^{N_v \times d_v} (d_v = 1152)$, and transforms it into a sequence of language-compatible embeddings, $g(f_v(\mathbf{v})) \in \mathbb{R}^{N_v \times d}$:

$$g(f_v(\mathbf{v})) = W_2 \cdot \mathrm{GELU}(W_1 \cdot f_v(\mathbf{v}) + b_1) + b_2,$$

where $W_1, W_2, b_1, b_2$ are learnable parameters. The projected visual embeddings are concatenated with the token embeddings of the textual prompt and jointly processed by the LLM decoder, enabling the decoder to interpret CBCT visual information within its native embedding space.

**LLM Decoder** For the novice configuration, we adopt Hulu-Med-4B as the language backbone. Text tokens derived from the prompt and projected visual tokens from the visual encoder are concatenated into a single multimodal sequence. The decoder then generates the full radiology report autoregressively, predicting each next token conditioned on all preceding text and visual tokens. This unified decoding formulation enables end-to-end report generation without introducing task-specific architectural modifications; adaptation to CBCT reporting is achieved via supervised fine-tuning.

### Clinical-Context Prompting

Clinical context was incorporated as textual conditioning to match deployable reporting workflows in which clinical diagnosis is typically available at interpretation time. For each CBCT study, the clinical diagnosis was authored by the consulting clinician during the patient encounter and then normalized to ICD-standardized diagnosis terms. We then inserted the diagnosis into a structured instruction prompt with an explicit field header to reduce ambiguity and stabilize model behavior across heterogeneous free-text inputs. A representative template as input is "Clinical Diagnosis: {*Clinical diagnosis*}. Provide a complete clinical CBCT report integrating findings and impression based on this 3D medical image".

### Instruction Tuning and Implementation Details

Instruction tuning was used to adapt Hulu-Med-4B to CBCT radiology report generation. Our instruction-tuning corpus contains 7,108 CBCT samples, each paired with a ground truth radiology report. To support bilingual reporting, we construct two language-specific instances per samples, resulting in 12,250 multimodal instruction–response samples in total. Each instance consists of a CBCT, an instruction prompt, and the corresponding full report target formatted with explicit section headers (“Findings:” and “Impression:”). The model is trained to generate the complete report autoregressively. To incorporate clinical context while discouraging over-reliance on diagnosis text, we define two prompt variants: (i) with clinical diagnosis conditioning using ICD-standardized diagnosis terms, and (ii) without diagnosis conditioning where the diagnosis field is omitted. We then constructed a fixed instruction-tuning dataset by sampling instances from these two variants by count, enforcing a 1:4 ratio (with diagnosis: without diagnosis) across the bilingual corpus. This prompt-mixing strategy reduces dependence on diagnosis conditioning and encourages robust report generation from imaging evidence.

**Image processing.** For each case, we used the reconstructed CBCT images at their original in-plane resolution and extracted an ordered slice sequence from the full scan range as the visual input. Prior to patch tokenization, images were intensity-normalized to reduce inter-scan variability. To control computational cost and maintain a consistent visual token budget, we sampled N axial slices per study using uniform sampling across the full scan range. Sampled slices were ordered by slice index and concatenated as a multi-image sequence (<image_1> … <image_N>). When scans contained fewer than N slices, we use all available slices; when scans exceeded N, we subsample uniformly. Here, we set $N = 96$.

**Training Details.** We performed full-parameter CBCT report-generation instruction tuning using PyTorch and DeepSpeed ZeRO-1. Training was conducted in bf16 with TF32 enabled, with a maximum context length of 16,384 tokens and up to 10,240 tokens allocated to multimodal inputs. We trained for 3 epochs on a Nvidia H200 with a per-device batch size of 2 and gradient accumulation to achieve an effective global batch size of 128. We optimized the model with AdamW and a cosine learning-rate schedule (warmup ratio 0.03; weight decay 0), using parameter-group learning rates of $1 \times 10^{-5}$ for the LLM decoder and multimodal projector and $2 \times 10^{-6}$ for the visual encoder. All parameters were initialized from Hulu-Med-4B pretrained weights, leveraging prior medical vision–language representations as a strong starting point for CBCT report-generation fine-tuning.

**Evaluation Framework**

To simulate a realistic clinical workflow, clinicians first conducted patient interviews and intraoral examinations to establish an initial diagnosis, followed by CBCT imaging. CBCTRepD was integrated directly into this workflow to generate draft reports conditioned on the CBCT volumes and the initial clinical diagnosis. These AI-generated drafts were then reviewed and edited by radiologists before being finalized and returned to the clinicians, faithfully reflecting the sequence of real-world report generation and review (Fig. 1b).

We recruited three tiers of radiologists representing distinct practice settings to comprehensively evaluate CBCTRepD across clinical scenarios. Residents (Novice) are graduate trainees in Oral and Maxillofacial Radiology programs, interpreting under senior supervision. Early-Career Board-Certified Radiologists (Intermediate) have 1–2 years of post-certification experience, serve as attending physicians for routine cases, and consult seniors for complex cases. Senior Board-Certified Radiologists (Senior) have more than five years of post-certification experience, act as full attending physicians responsible for final diagnoses and supervision (Fig. 1d; Fig. S1).

For each case comprising CBCT images and the initial clinical diagnosis, CBCTRepD generated an AI draft report, while the three radiologist cohorts independently produced manual reports, resulting in three sets of collaboration reports (AI+Novice, AI+Intermediate, AI+Senior). Additionally, to assess actual contribution of CBCTRepD to report quality, separate radiologists from each tier wrote manual reports without AI assistance. This procedure yielded a total of seven reports under identical case conditions.

Finally, clinical performance was assessed through a multi-faceted evaluation framework incorporating linguistic quality, radiologist-centered assessment, and clinician-centered evaluation. This approach allowed comprehensive measurement of both report-writing capability and the practical clinical value of the reports from multiple perspectives.

**Evaluation of Generation Reports**

Automatic evaluation encompassed both linguistic- and diagnosis-level metrics. Linguistic quality was quantified using BLEU, ROUGE-L, METEOR, and BERTScore, reported separately for Chinese and English. Clinical content correctness was assessed by comparing extracted impression entities from model outputs with expert references, with both Accuracy and Recall reported (Fig. 1d).

Radiologist-centered evaluation included preference ranking, quality score, and error type annotation. Preference ranking assigned ranks within a fixed candidate set, ranging from 1 (best) to 4 (worst), across AI-generated reports and manual reports produced by three tiers of radiologists (Fig. 4d; Table S3). Quality score followed a four-dimensional rubric—factual consistency, coherence, medical safety, and clinical utility—rated on a four-point scale (Table S9). Radiologists then annotated omissions and incorrect entries for Findings and Impressions, recording each type's count and clinical significance (Fig. 4d; Tables S4–S5).

From the initial clinical diagnoses of four subspecialty departments, 25 reports per department were randomly selected from a total of 300 reports. Clinician subspecialists from each department reviewed these reports, providing preference rankings and quality scores from a clinical perspective (Table S3).

**Evaluation of CBCTRepD for Three-Tier Radiologist Support**

Collaboration reports were evaluated using the same framework of preference ranking, quality scores, and error-type annotation, allowing assessment of CBCTRepD's impact on each tier of radiologist support (Fig. 1d; Fig. 5).

Preference ranking was conducted within a fixed candidate set excluding AI-only outputs, with ranks from 1 (best) to 6 (worst) across manual and collaboration reports (Fig. 1d; Fig. 5b; Table S6). Quality scores followed the same four-dimensional rubric (Table S9) and were summarized as mean scores and Δ relative to paired manual reports (Δscore = Collaboration − Manual) (Fig. 5c). Error-type annotation quantified omissions and incorrect entries in Findings and Impressions, recording counts and clinical significance. Δ relative to paired manual reports was calculated to evaluate the impact of collaboration (Fig. 5d; Tables S7–S8). (Fig. 5d; Tables S7–S8). Clinician evaluation similarly compared collaboration and manual reports from a clinical perspective.

**Statistical analysis**

Differences across report groups were analyzed using non-parametric tests suitable for ordinal outcomes and non-normal error distributions. For multi-group comparisons within each subspecialty stratum, two-sided Kruskal–Wallis tests were applied, with Holm-adjusted P values used for post hoc pairwise comparisons where applicable.

**Ethics statement**

This study was approved by the local Institutional Review Board (ZJUSSIRB-2025-191). All imaging data and reports were de-identified prior to analysis. The need for informed consent was waived due to the retrospective nature of the study.

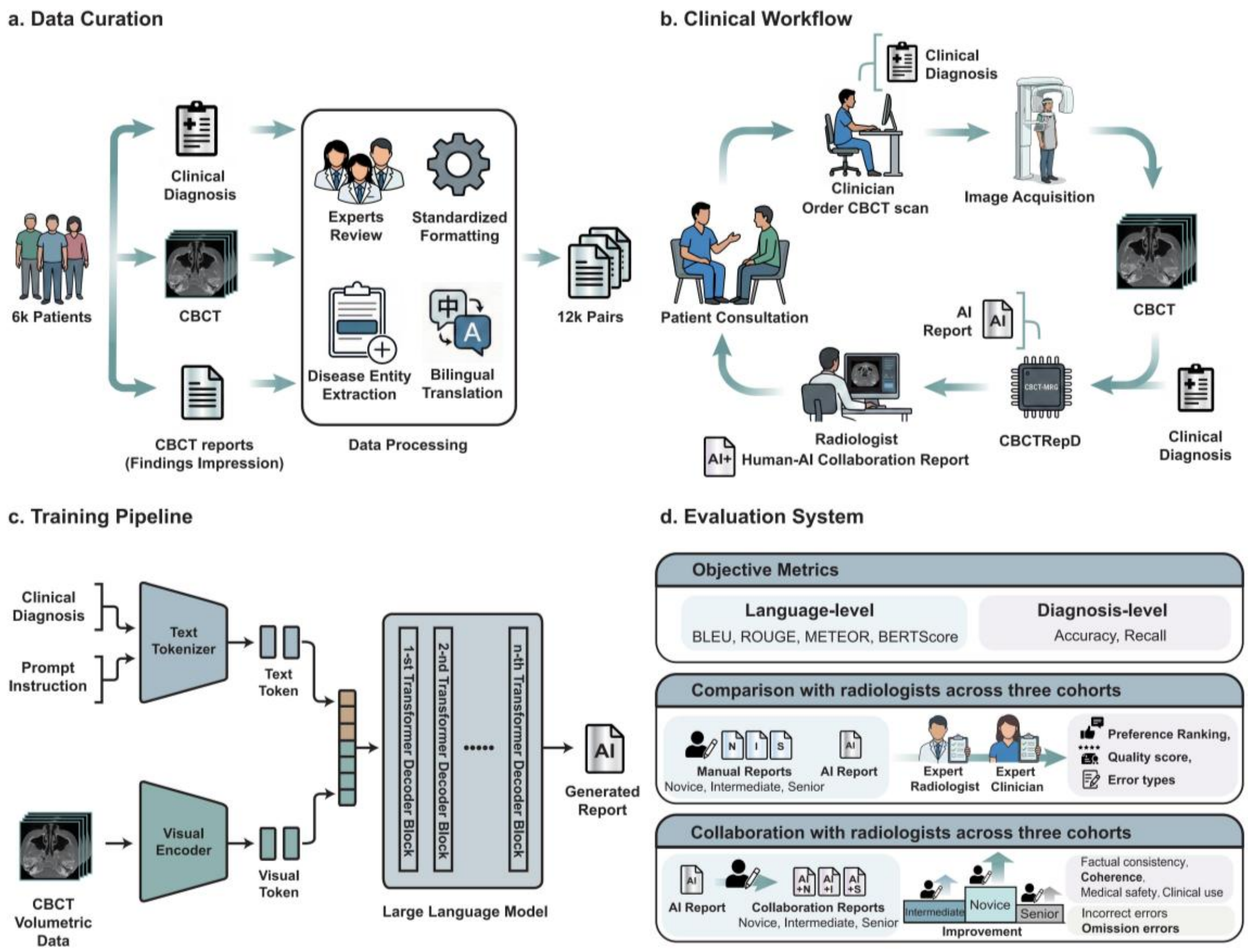

**Figure 1. Overall framework of CBCTRepD and study design.**

**a, Data curation.** CBCT scans, initial clinical diagnoses, and corresponding radiology reports were collected from 7,303 patients. The reports were curated through expert review, standardized formatting, disease-entity extraction, and bilingual translation, yielding 14,016 CBCT–report pairs for model development.

**b, Clinical workflow.** b, Clinical workflow. In routine practice, CBCT is acquired and processed by CBCTRepD together with the intial clinical diagnosis to generate a draft radiology report. The draft is then refined by radiologists and delivered to the clinician for patient consultation.

**c, Training pipeline.** A vision–language architecture is trained to generate CBCT radiology reports. A visual encoder converts CBCT volumetric data into visual tokens, while a text tokenizer encodes clinical diagnoses and task instructions into text tokens. These multimodal tokens are jointly modeled by a transformer-based large language model to generate the final report.

**d, Evaluation system.** Model outputs were evaluated from three perspectives: (i) objective metrics; (ii) comparison with radiologists at three experience levels, in which clinical and radiological experts compared AI-generated reports with manual reports written by novice, intermediate, and senior radiologists using preference ranking, quality scores, and structured error categorization for the Findings and Impression sections; and (iii) human–AI collaboration with radiologists at three experience levels, in which collaboration reports were compared with manually written reports from radiologists at the same three tiers.

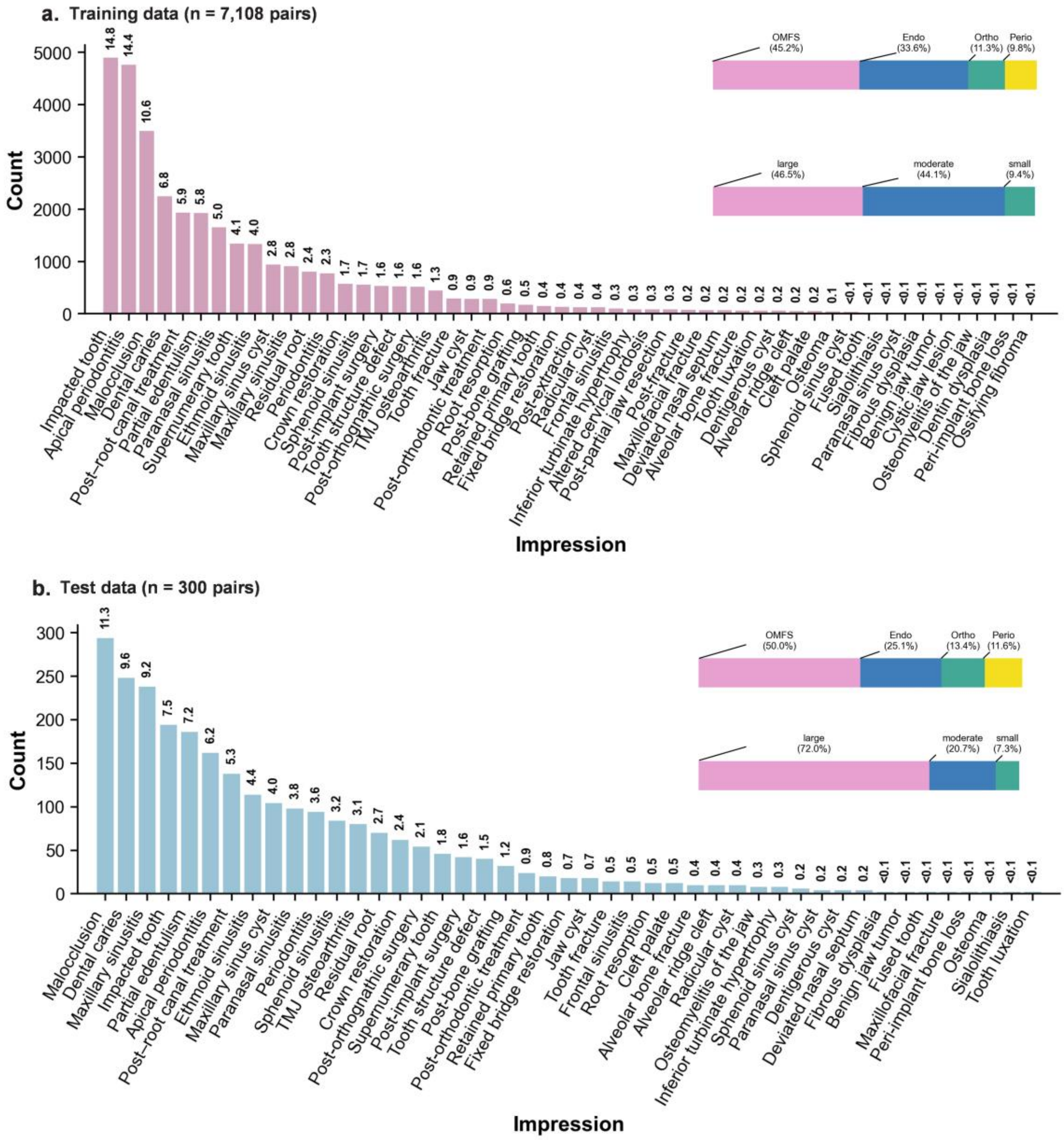

**Figure 2. Data statistics for training and test cohorts.**

**a,** Distribution of diagnostic impressions in the training dataset. Bars indicate the count of pairs annotated with each impression term, ordered from most to least frequent along the x-axis.

**b,** Corresponding impression distribution in the independent test dataset. For each cohort, the stacked horizontal bars (right) summarize the composition of cases by clinical discipline—oral and maxillofacial surgery (OMFS), endodontics (Endo), orthodontics (Ortho) and periodontology, implantology, and prosthodontics (Perio-Implant-Prostho)—and by CBCT scanning field-of-view (FOV) category (large, moderate, small). Segment labels report the percentage of pairs within the respective cohort.

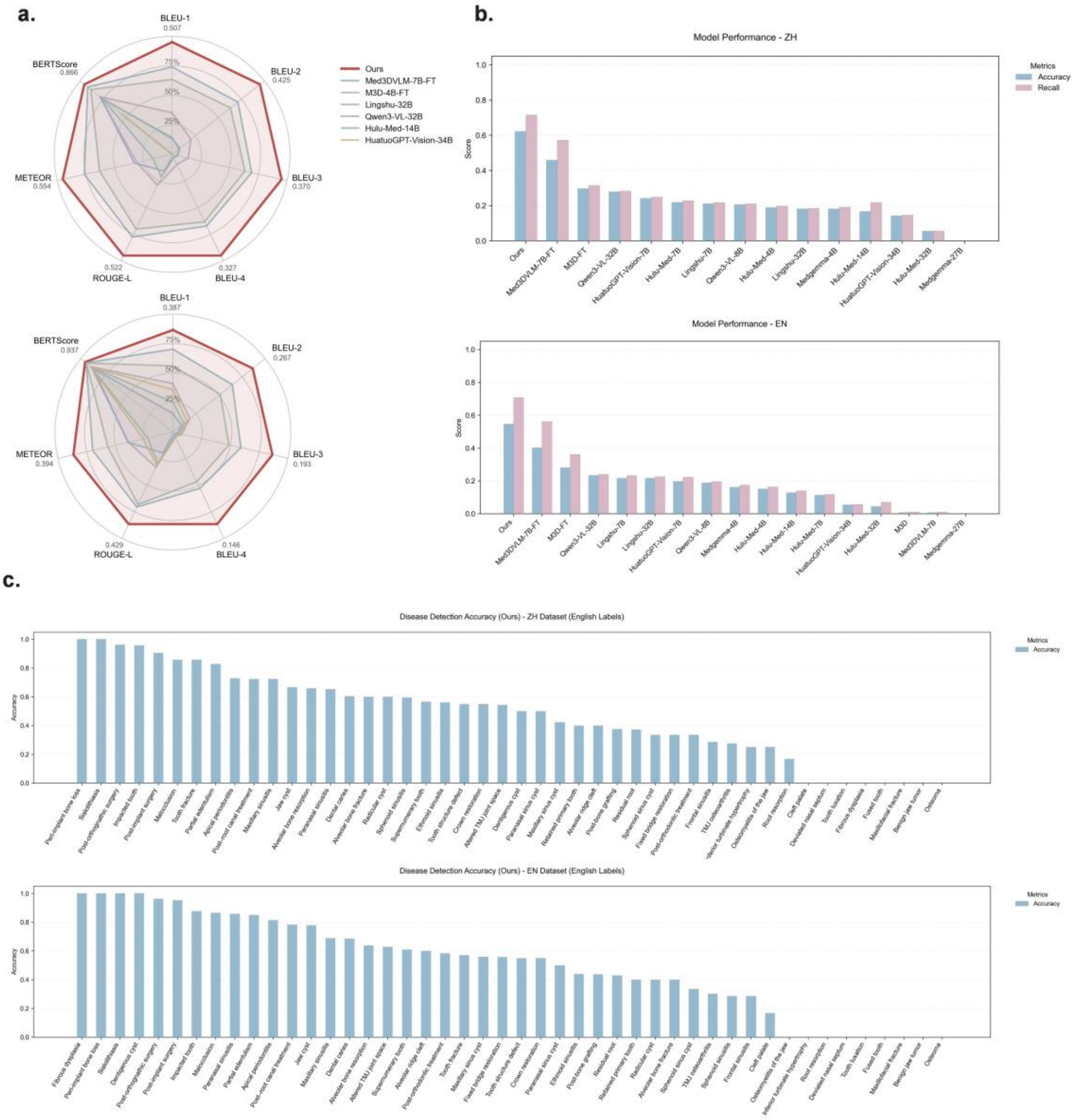

**Figure 3. Linguistic and Diagnostic Performance of the Proposed Method.**

a, Radar plots showing automatic natural language generation metrics for Chinese (ZH, top) and English (EN, bottom) CBCT reports, including BLEU, ROUGE-L, METEOR, and BERTScore.

b, Overall disease-level accuracy and recall in ZH (top) and EN (bottom) for each baseline and the proposed method.

c, Per-disease detection accuracy of the proposed method on the ZH (top) and EN (bottom) test sets, with diseases ordered by accuracy.

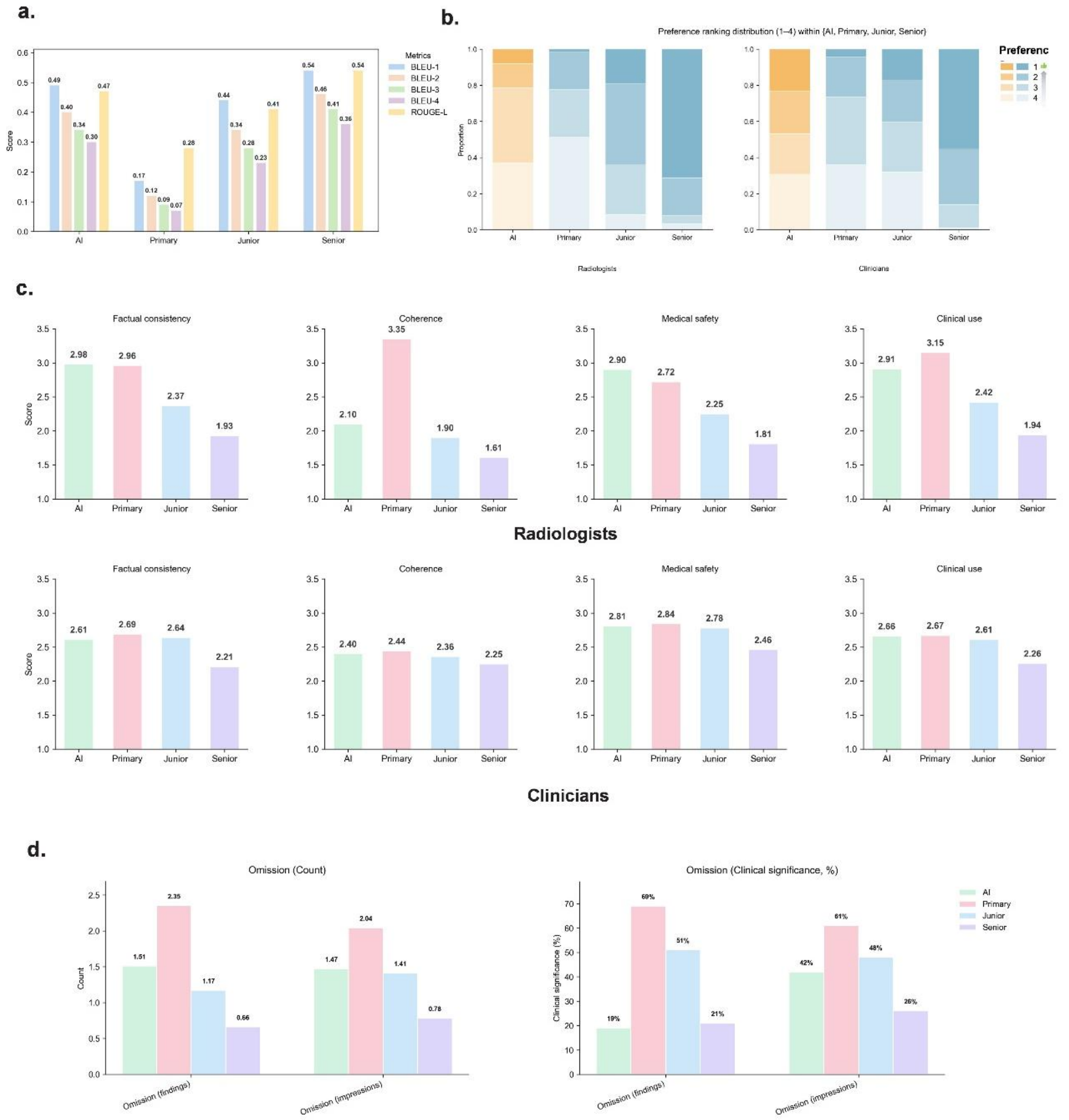

**Figure 4. Comparative Performance of AI Reports and Manual Reports by Community, Intermediate, and Senior Radiologists.**

**a**, Linguistic Performance. Automatic report-generation performance for AI, Novice, Intermediate and Senior reports quantified using BLEU-1/2/3/4 and ROUGE-L. Values above bars indicate the corresponding metric scores.

**b**, Preference Ranking. Expert preference distributions shown as stacked proportions of rank assignments (Rank 1 = most preferred; Rank 4 = least preferred) for radiologists (left) and clinicians (right) when comparing report sources (AI vs. human-written reports by seniority).

**c**, Quality scores. Mean subjective ratings across four dimensions—factual consistency, coherence, medical safety, and clinical usefulness—reported separately for radiologists (top row) and clinicians (bottom row).

d, Omission analysis. Mean omission burden in Findings and Impressions shown as omission count (left) and the proportion of clinically significant omissions (right) across AI, Novice, Intermediate and Senior reports.

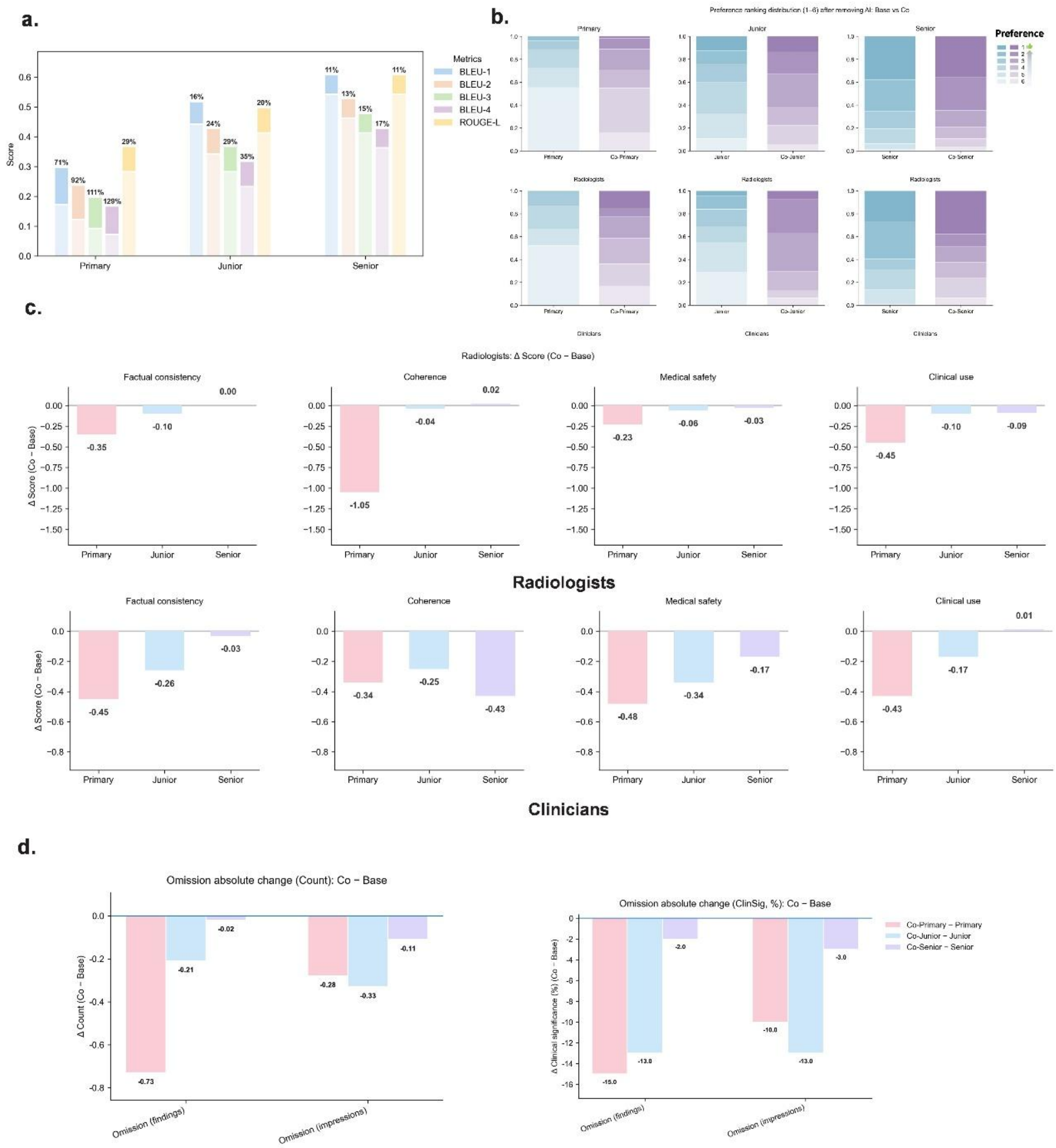

**Figure 5. Performance Improvements of Human-AI Collaboration Compared to Manual Reporting by Radiologists**

**a**, Linguistic Performance improvements. Relative change (percentage) in automatic NLG metrics (BLEU-1/2/3/4 and ROUGE-L) for collaboration reports compared with their corresponding manual baselines (Co-Novice vs Novice, Co-Intermediate vs Intermediate, Co-Senior vs Senior); percentages denote the relative improvement.

**b,** Preference shift. Preference ranking distributions shown as stacked proportions of rank assignments (Rank 1 = most preferred; Rank 6 = least preferred) for radiologists (top row) and clinicians (bottom row) when comparing each manual baseline against its collaboration counterpart after excluding AI-only reports.

**c**, Quality-score improvements. Mean change in subjective quality ratings (Δ score= collaborative-baseline) across four dimensions, factual consistency, coherence, medical safety, and clinical usefulness, reported separately for radiologists (top) and clinicians (bottom).

Negative values indicate higher ratings for collaboration reports.

**d**, Omission reduction. Change in omission burden (Δ = Co-Base) for Findings and Impressions, shown as absolute change in omission count (left) and change in the proportion of clinically significant omissions (right).

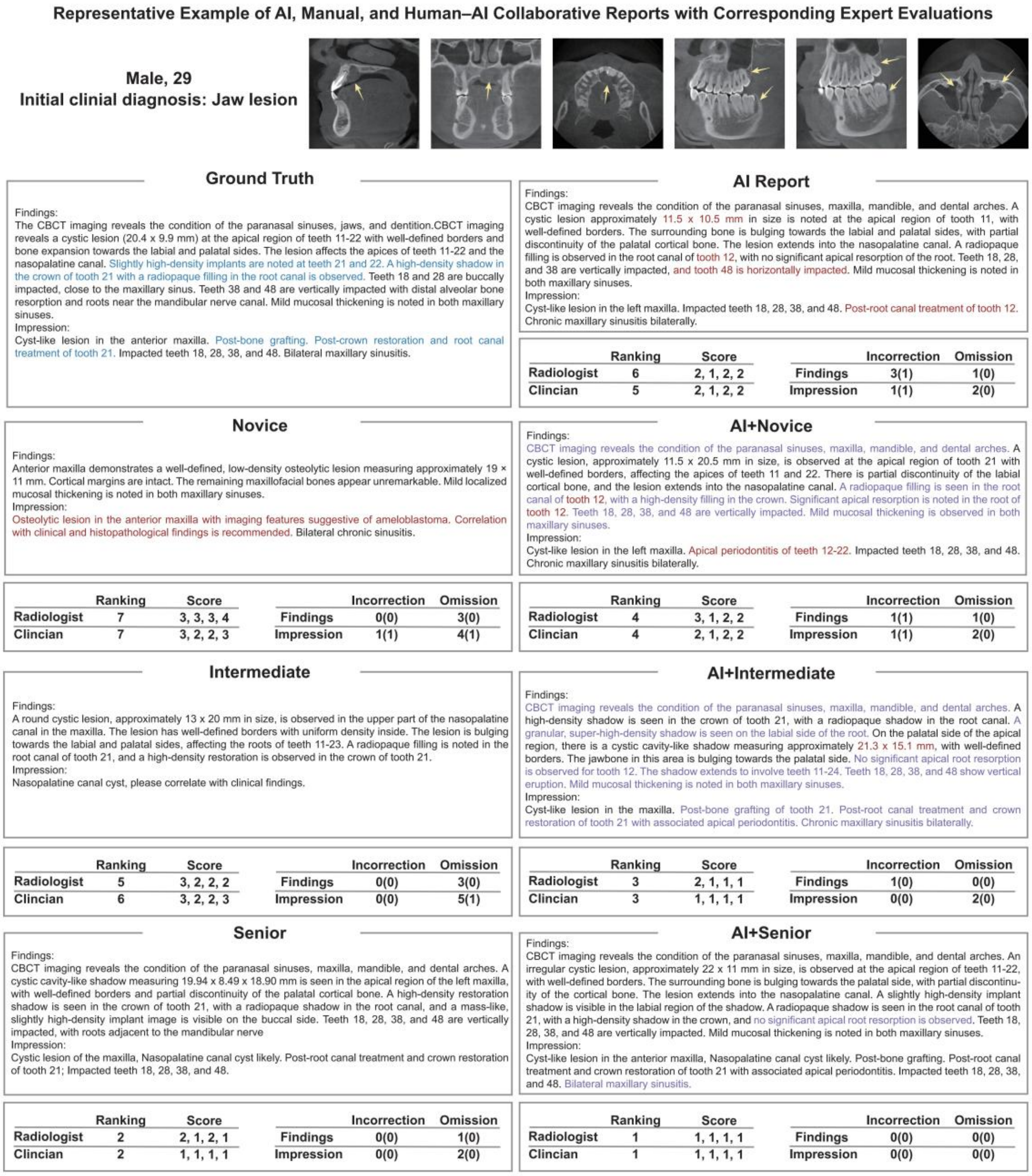

**Fig. 6 Representative Example of AI, Manual, and Human–AI Collaborative Reports with Corresponding Expert Evaluations**

Yellow arrows indicate the specific locations of lesions. Blue text marks findings or impressions omitted from the AI-generated report. Red text indicates errors in the findings or impression relative to the ground truth. Purple text highlights additional information in the AI+Human report relative to the manual report, illustrating the assistive contribution of the AI system. Radiologist and clinician ratings are shown below each report. Ranking denotes the preference order, and Score denotes quality ratings across four dimensions: factual consistency, coherence, medical safety and clinical usefulness.

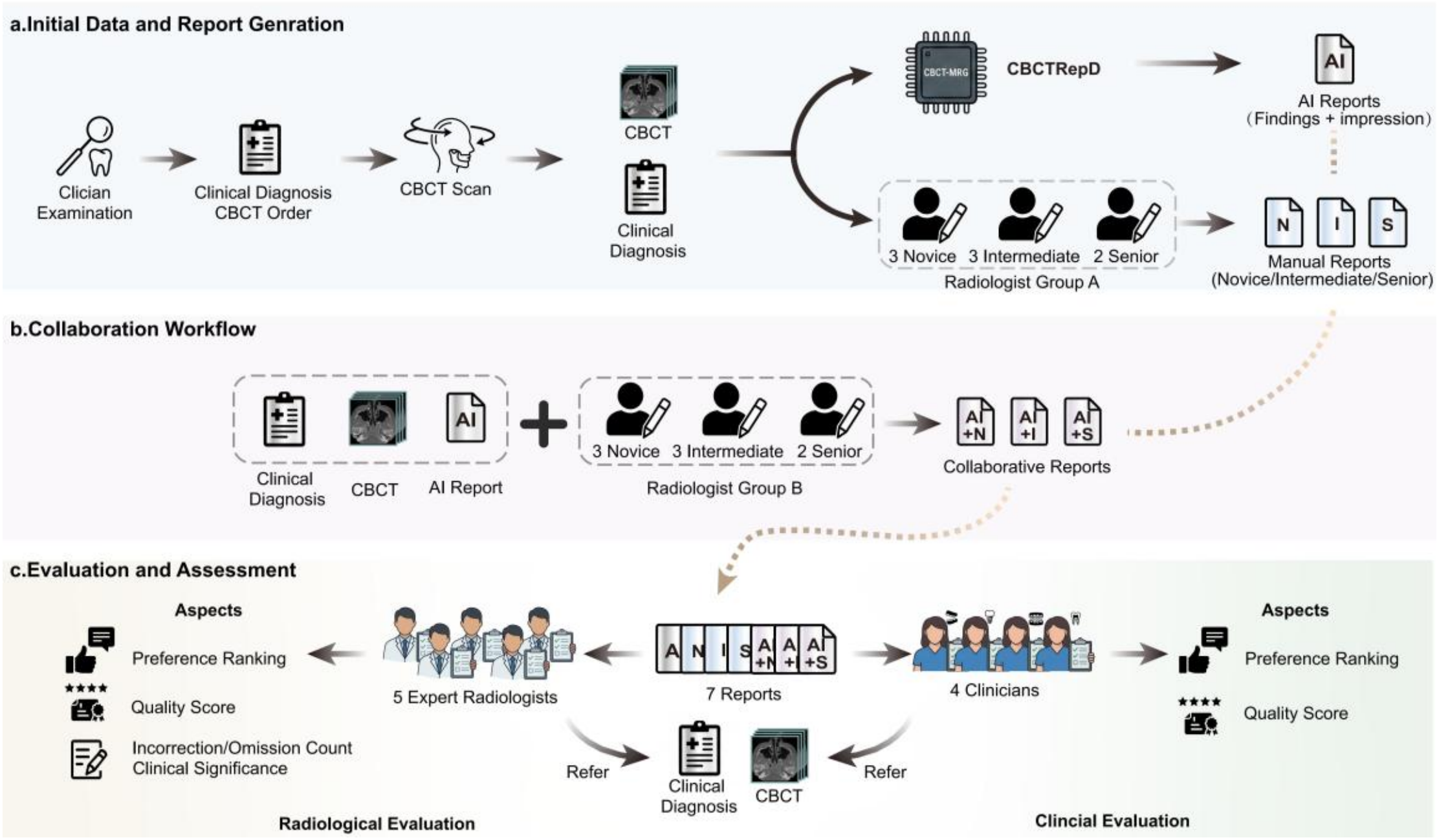

**Figure S1. Overview of the Clinical Workflow for Report Generation, Collaboration, and Evaluation**

**a,** The initial data collection begins with clinician examination, clinical diagnosis, and CBCT scan, which serve as the foundation for Generation Reports through the CBCTRepD system. Additionally, three groups of radiologists (Novice, Intermediate, and Senior) independently produce manual reports, representing purely radiologist-generated outputs.

**b,** Based on the AI Reports, CBCT data, and clinical diagnosis, the other radiologists from three groups (Novice, Intermediate, and Senior) generate Collaborative Reports by revising the AI reports seperately.

**c,** The evaluation process involves collecting the seven reports along with the corresponding CBCT scans and initial clinical diagnosis. On the left side, Expert Radiologists evaluate all reports according to various evaluation aspects listed. Additionally, 4 clinicians from 4 departments: Oral and Maxillofacial Surgery (OMFS), Endodontics (Endo), Orthodontics (Ortho), and Periodontology, Implantology, and Prosthodontics (Perio-Implant-Prostho) assessed the reports from clinical aspects.

## Example of Bilingual Generation Report task in different models

**Male, 44**
**Clinial diagnosis: Partial Edentulism**

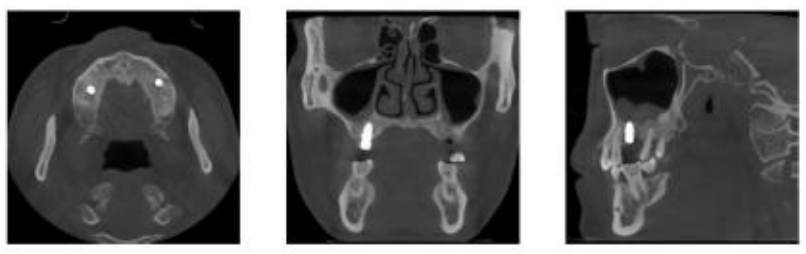

### Ground Truth

影像所见：该影像呈现了副鼻窦、上下颌骨及牙列的情况。16、26牙缺失，对应区域各见一枚种植体植入。17、27根管内见阻射影，17远中冠部低密度影。两侧上颌窦及部分筛窦黏膜增厚。
诊断印象：牙列缺损(16，26)：种植术后：17，27根管治疗术后：17牙体缺损：两侧上颌窦及筛窦炎。

Findings:
The imaging reveals the condition of the paranasal sinuses, maxilla, mandible, and dental arches. Teeth 16 and 26 are missing, with implants placed in the corresponding areas. Radiopaque fillings are observed in the root canals of teeth 17 and 27, with a low-density shadow in the distal crown of tooth 17. Mild mucosal thickening is noted in both maxillary sinuses and part of the ethmoid sinuses.

Impression:
Partial edentulism (teeth 16, 26); post-implant surgery; post-root canal treatment of teeth 17 and 27; tooth structure defect in tooth 17; bilateral maxillary sinusitis and ethmoid sinusitis.

### Ours

影像所见：CBCT影像呈现了副鼻窦、上下颌骨及牙列的情况。16、26牙缺失，对应区域各见一枚种植体植入。17、27牙根管内见阻射影，根尖周未见明显阴影。18、28垂直阻生。两侧上颌窦部分粘膜稍增厚。
诊断印象：牙列缺损(16，26)：种植术后：17，27根管治疗术后：18，28阻生齿：两侧上颌窦慢性炎症。

Findings:
CBCT imaging reveals the condition of the paranasal sinuses, maxilla, mandible, and dental arches. Teeth 16 and 26 are missing, with implants placed in the corresponding areas. Radiopaque fillings are observed in the root canals of teeth 17 and 27, with no significant shadow in the periapical region. Teeth 18 and 28 are vertically impacted. Mild mucosal thickening is noted in both maxillary sinuses.

Impression:
Partial edentulism (teeth 16, 26); post-implant surgery; post-root canal treatment of teeth 17 and 27; impacted teeth 18 and 28; bilateral chronic maxillary sinusitis.

### Med3DVLM-7B-FT

影像所见：上下颌牙列缺失，左侧上颌骨后牙区见4枚种植体植入。右侧上颌窦内见软组织密度影。
诊断印象：牙列缺损：种植术后

Findings:
Both the upper and lower dental arches are edentulous. Four implants are placed in the posterior region of the left maxilla. A soft tissue density shadow is observed in the right maxillary sinus.

Impression:
Partial edentulism; post-implant surgery.

### M3D-FT

影像所见：11、21牙缺失，对应牙槽骨吸收变窄，高度降低，骨纹理稍稀疏。

Findings:
Teeth 11 and 21 are missing, with corresponding alveolar bone resorption, narrowing, and reduced height. The bone trabeculation appears slightly diminished.

### Lingshu-32B

影像表现：CBCT图像显示上颌和下颌的牙槽骨结构。上颌中可见多个牙齿缺失，尤其是前牙区和磨牙区。下颌也存在多颗牙齿缺失，主要集中在前牙区。牙槽骨的高度和宽度在多个区域有所减少，表明骨质吸收。上颌窦区域清晰可见，没有明显的窦腔病变。影像中未发现明显的病理性骨改变或异常结构。
诊断印象：根据影像表现，患者存在多颗牙齿缺失，主要集中在上颌和下颌的前牙区和磨牙区。牙槽骨吸收明显，可能需要进行种植牙手术或其他修复治疗。建议进行进一步的临床评估和治疗计划制定。

Findings:
CBCT images show the alveolar bone structure of both the maxilla and mandible. Multiple teeth are missing in the maxilla, particularly in the anterior and molar regions. The mandible also shows several missing teeth, mainly concentrated in the anterior region. The height and width of the alveolar bone are reduced in several areas, indicating bone resorption. The maxillary sinus region is clearly visible with no significant sinus cavity pathology. No obvious pathological bone changes or abnormal structures are identified in the images.

Impression:
Based on the imaging findings, the patient has multiple missing teeth, primarily in the anterior and molar regions of both the maxilla and mandible. Significant alveolar bone resorption is present, which may require dental implant surgery or other restorative treatments. Further clinical evaluation and treatment planning are recommended.

### HuatuoGPT-Vision-7B

影像学检查：
检查类型：锥形束计算机断层扫描（CBCT）
扫描范围：上颌骨及下颌骨区域
扫描参数：具体参数需根据设备记录提供
影像学发现：
1. 牙列情况：上颌骨及下颌骨区域可见明显的牙列缺损。缺失牙齿部位的牙槽嵴高度和宽度明显降低，部分区域牙槽骨吸收较为严重。部分剩余牙齿存在不同程度的牙周支持组织丧失。
2. 牙槽骨状况：
牙槽骨密度普遍减低，提示骨质疏松或骨量不足。在缺牙区周围，牙槽骨边缘不规则，部分区域骨皮质变薄。下颌骨后部及上颌骨前部的牙槽骨吸收更为显著，可能与长期缺失牙齿有关。
3. 邻近解剖结构：
上颌窦底壁在部分缺牙区显示清晰，未见明显炎症或病变迹象。下颌神经管在下颌骨体部显示完整，但靠近下颌支区域骨质稍显稀疏。
4. 软组织观察：
软组织形态基本正常，未见明显肿胀或异常增生。口腔黏膜未见溃疡、肿物等病理性改变。
诊断印象：
主要诊断：牙列缺损（全口或局部）。次要诊断：牙槽骨吸收（重度）。相关诊断：骨量不足导致种植修复困难；可能存在牙周炎史。

Imaging Examination:
Examination Type: Cone Beam Computed Tomography (CBCT)
Scan Area: Maxilla and mandible
Scan Parameters: Specific parameters to be provided based on equipment records

Findings:
1. Dental Status: Significant edentulism is noted in both the maxilla and mandible. The alveolar ridge height and width are notably reduced in the missing tooth areas, with severe alveolar bone resorption in some regions. Some remaining teeth show varying degrees of periodontal support loss.
2. Alveolar Bone Condition: Generalized decrease in alveolar bone density suggests osteoporosis or insufficient bone volume. Irregular alveolar bone margins are seen in edentulous areas, with some regions showing cortical bone thinning. Significant alveolar bone resorption is noted in the posterior mandible and anterior maxilla, likely related to long-term tooth loss.
3. Adjacent Anatomical Structures: The floor of the maxillary sinus is clearly visible in areas of tooth loss, with no signs of inflammation or pathology. The mandibular nerve canal appears intact in the body of the mandible, with slightly reduced bone density near the mandibular ramus.
4. Soft Tissue Observation: Soft tissue morphology is generally normal, with no significant swelling or abnormal growth. No pathological changes such as ulcers or masses are seen in the oral mucosa.

Impression:
Primary Diagnosis: Dental edentulism (full or partial)
Secondary Diagnosis: Severe alveolar bone resorption
Related Diagnosis: Insufficient bone volume, which may complicate implant restoration; possible history of periodontitis.

**Figure S2. Example of Bilingual Generation Report task in different models**

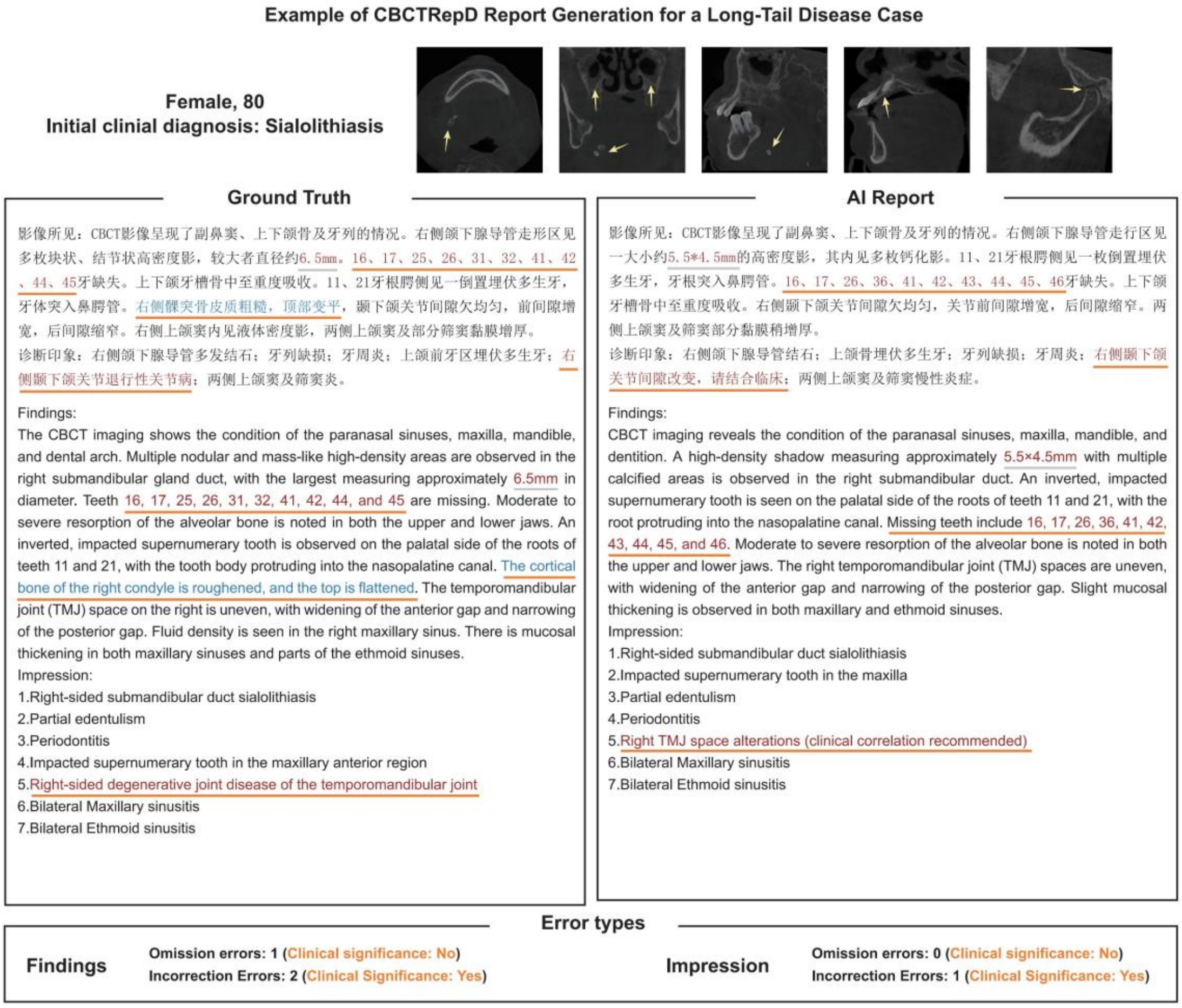

**Figure S3. Example of CBCTRepD Report Generation for a Long-Tail Disease Case**

Yellow arrows indicate the specific locations of lesions. Blue text marks findings or impressions omitted from the AI-generated report. Red text indicates errors in the findings or impression relative to the ground truth. Orange underlining denotes errors considered clinically significant.

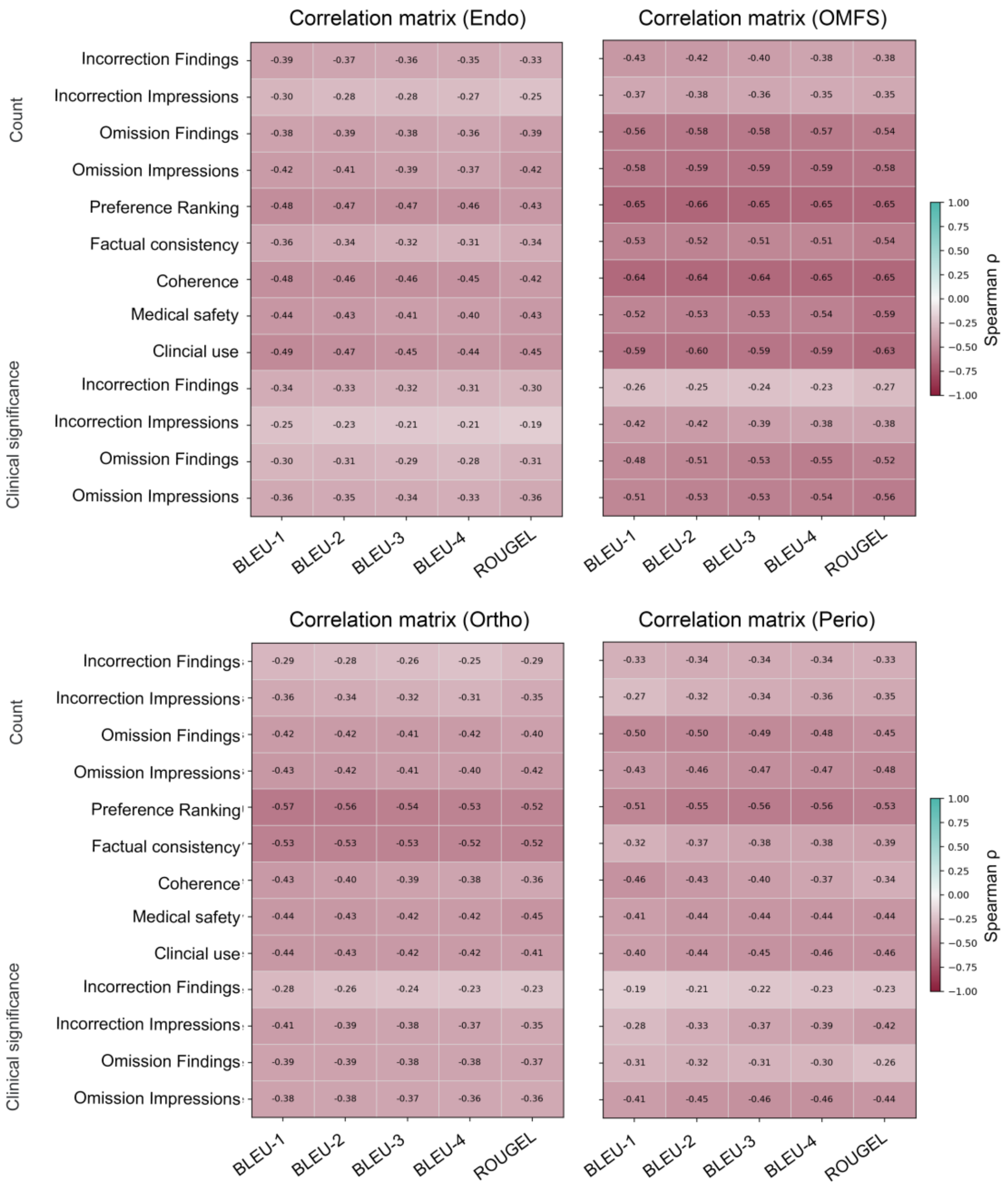

**Figure S4. Correlation of Quality Scores and NLG Metrics Across Four Subspecialties in Radiologists**

This figure presents the Spearman correlation matrices showing the relationship between quality scores and natural language generation (NLG) metrics for four subspecialties: Endo, OMFS, Ortho, Perio-Implant-Prostho. The quality scores assess Incorrection Findings, Incorrection Impressions, Omission Findings, Omission Impressions, Preference Ranking, Factual Consistency, Coherence, Medical Safety, and Clinical Use, while the NLG metrics include BLEU-1, BLEU-2, BLEU-3, BLEU-4, and ROUGE-L. Each matrix illustrates the Spearman correlation coefficients (ρ) between these quality score dimensions and the NLG metrics. Correlation strength is visualized using a color scale, where darker red shades represent stronger negative correlations, while green shades represent stronger positive correlations.

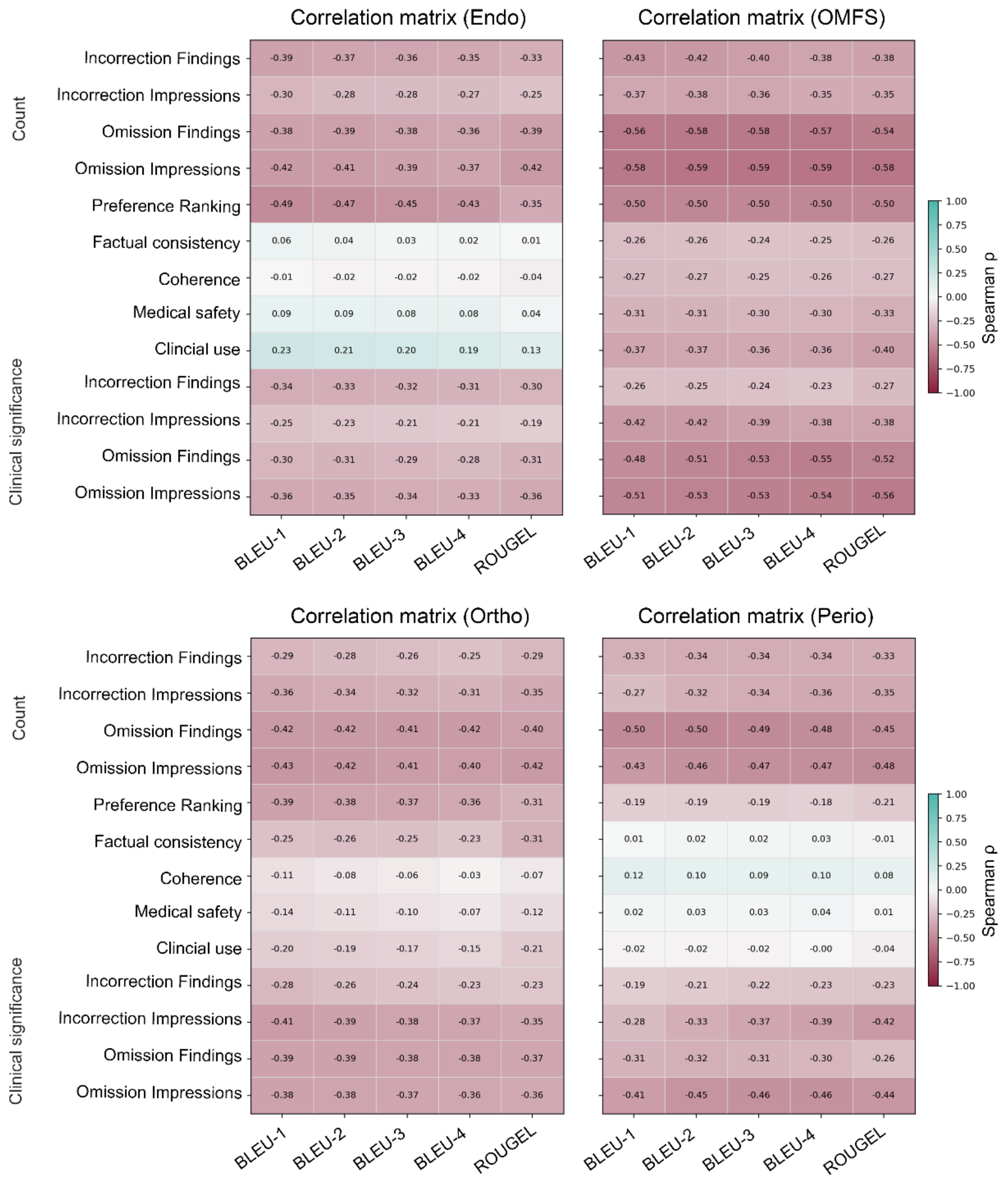

**Figure S5. Correlation of Quality Scores and NLG Metrics Across Four Subspecialties in Clinicians**

This figure presents the Spearman correlation matrices illustrating the relationship between quality scores and natural language generation (NLG) metrics for four subspecialties: Endo, OMFS, Ortho, Perio-Implant-Prostho. The quality scores include Incorrection Findings, Incorrection Impressions, Omission Findings, Omission Impressions, Preference Ranking, Factual Consistency, Coherence, Medical Safety, and Clinical Use, while the NLG metrics are BLEU-1, BLEU-2, BLEU-3, BLEU-4, and ROUGE-L. The matrices show the Spearman correlation coefficients (ρ) between the quality scores and the NLG metrics, with color shading indicating the strength of the correlation, where darker red shades represent stronger negative correlations and green shades represent stronger positive correlations.

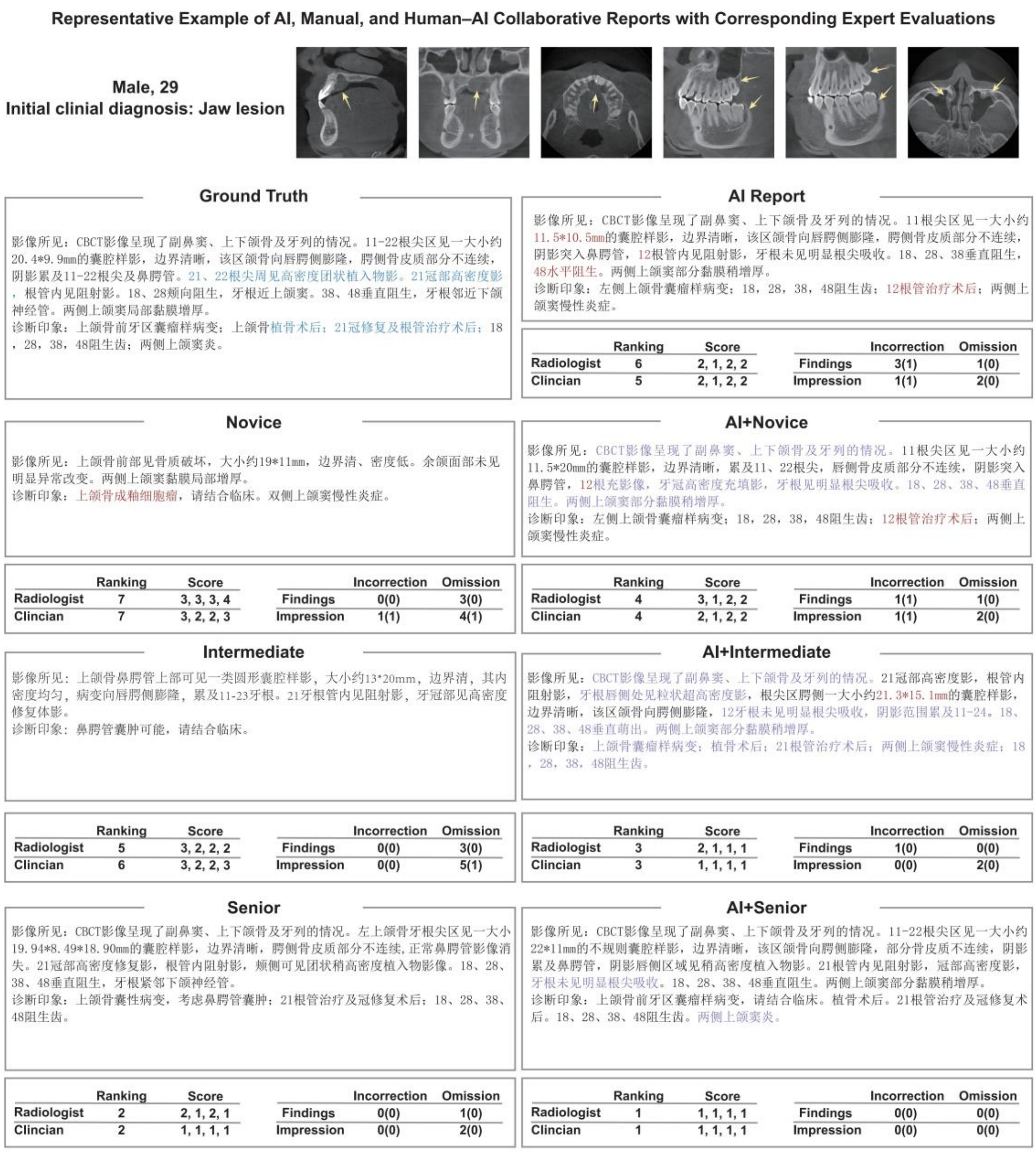

**Figure S6. Representative Example of AI, Manual, and Human–AI Collaborative Reports with Corresponding Expert Evaluations (Chinese)**

Yellow arrows indicate the specific locations of lesions. Blue text marks findings or impressions omitted from the AI-generated report. Red text indicates errors in the findings or impression relative to the ground truth. Purple text highlights additional information in the AI+Human report relative to the manual report, illustrating the assistive contribution of the AI system. Radiologist and clinician ratings are shown below each report. Ranking denotes the preference order, and Score denotes quality ratings across four dimensions: factual consistency, coherence, medical safety and clinical usefulness.

**Table S1. Demographic characteristics and CBCT acquisition parameters of dataset**

| Characteristics | Training | Test |
|---|---|---|
| **Sex** | | |
| Female | 4008 | 166 |
| Male | 3099 | 134 |
| **Age (years)** | 26.9±17.1 | 26.93±17.07 |
| **Field-of-View** | | |
| Large | 3307 | 216 |
| Moderate | 3132 | 62 |
| Small | 668 | 22 |
| **PixelSpacing** | | |
| [0.3, 0.3] | 5522 | 239 |
| [0.25, 0.25] | 525 | 43 |
| [0.2, 0.2] | 77 | 6 |
| [0.16, 0.16] | 30 | 5 |
| [0.15, 0.15] | 565 | 7 |
| [0.125, 0.125] | 388 | - |
| **KV** | | |
| 110 | 6017 | 245 |
| 90 | 1061 | 49 |
| 100 | 29 | 6 |
| **Device** | | |
| NEWTOM VGIEVO | 6017 | 245 |
| Morita | 925 | 49 |
| Planmeca ProMax 3DMid | 89 | 6 |

Table S2. Quantitative performance of CBCT report generation on different baselines

| Method | BLEU-1 | BLEU-2 | BLEU-3 | BLEU-4 | ROUGE-L | METEOR | BERTScore | Recall |
|---|---|---|---|---|---|---|---|---|
| ZH | | | | | | | | |
| Qwen3-VL-8B | 0.066 | 0.029 | 0.014 | 0.006 | 0.075 | 0.142 | 0.644 | 0.211 |
| Qwen3-VL-32B | 0.068 | 0.036 | 0.018 | 0.008 | 0.082 | 0.182 | 0.671 | 0.283 |
| Medgemma-4B | 0.073 | 0.023 | 0.009 | 0.004 | 0.077 | 0.115 | 0.605 | 0.192 |
| Lingshu-32B | 0.179 | 0.087 | 0.052 | 0.031 | 0.151 | 0.170 | 0.680 | 0.185 |
| Hulu-Med-7B | 0.025 | 0.011 | 0.007 | 0.004 | 0.076 | 0.057 | 0.639 | 0.228 |
| Hulu-Med-4B | 0.079 | 0.036 | 0.019 | 0.012 | 0.104 | 0.085 | 0.650 | 0.198 |
| Hulu-Med-32B | 0.015 | 0.007 | 0.004 | 0.003 | 0.027 | 0.021 | 0.532 | 0.056 |
| Hulu-Med-14B | 0.072 | 0.031 | 0.017 | 0.010 | 0.111 | 0.091 | 0.655 | 0.218 |
| HuatuoGPT-Vision-7B | 0.106 | 0.049 | 0.023 | 0.010 | 0.111 | 0.184 | 0.662 | 0.249 |
| HuatuoGPT-Vision-34B | 0.002 | 0.000 | 0.000 | 0.000 | 0.002 | 0.001 | 0.476 | 0.146 |
| M3D-4B-FT | 0.322 | 0.269 | 0.234 | 0.207 | 0.367 | 0.350 | 0.764 | 0.315 |
| Med3DVLM-7B-FT | 0.376 | 0.303 | 0.256 | 0.220 | 0.406 | 0.423 | 0.793 | 0.572 |
| Ours | **0.483** | **0.405** | **0.352** | **0.311** | **0.497** | **0.528** | **0.825** | **0.715** |
| EN | | | | | | | | |
| Qwen3-VL-8B | 0.060 | 0.022 | 0.007 | 0.003 | 0.079 | 0.121 | 0.791 | 0.196 |
| Qwen3-VL-32B | 0.061 | 0.025 | 0.008 | 0.003 | 0.084 | 0.154 | 0.805 | 0.240 |
| Medgemma-4B | 0.074 | 0.025 | 0.007 | 0.003 | 0.088 | 0.113 | 0.792 | 0.174 |
| Lingshu-7B | 0.160 | 0.051 | 0.017 | 0.008 | 0.137 | 0.146 | 0.831 | 0.232 |
| Lingshu-32B | 0.159 | 0.049 | 0.015 | 0.007 | 0.142 | 0.152 | 0.832 | 0.225 |
| Hulu-Med-7B | 0.043 | 0.014 | 0.005 | 0.003 | 0.096 | 0.053 | 0.826 | 0.118 |
| Hulu-Med-4B | 0.051 | 0.018 | 0.007 | 0.004 | 0.109 | 0.063 | 0.830 | 0.163 |
| Hulu-Med-32B | 0.099 | 0.038 | 0.013 | 0.006 | 0.115 | 0.091 | 0.816 | 0.070 |
| **Hulu-Med-14B** | 0.098 | 0.032 | 0.011 | 0.006 | 0.126 | 0.087 | 0.833 | 0.140 |
| **HuatuoGPT-Vision-7B** | 0.109 | 0.038 | 0.012 | 0.006 | 0.108 | 0.141 | 0.813 | 0.222 |
| **HuatuoGPT-Vision-34B** | 0.139 | 0.040 | 0.014 | 0.007 | 0.134 | 0.102 | 0.831 | 0.057 |
| **M3D-4B-FT** | 0.217 | 0.137 | 0.094 | 0.068 | 0.292 | 0.223 | 0.879 | 0.361 |
| **Med3DVLM-7B-FT** | 0.272 | 0.172 | 0.114 | 0.077 | 0.302 | 0.275 | 0.881 | 0.562 |
| **Ours** | **0.335** | **0.231** | **0.167** | **0.126** | **0.371** | **0.341** | **0.892** | **0.708** |

**Table S3. Detection accuracy of our method for the Impression**

| Impression | ZH | EN | Impression | ZH | EN |
|---|---|---|---|---|---|
| **Malocclusion** | 0.857 | 0.864 | **Fixed dental bridge** | 0.333 | 0.556 |
| **Dental caries** | 0.605 | 0.685 | **Jaw cyst** | 0.667 | 0.778 |
| **Maxillary sinusitis** | 0.723 | 0.689 | **Tooth fracture** | 0.857 | 0.571 |
| **TMJ joint space change** | 0.543 | 0.629 | **Frontal sinusitis** | 0.286 | 0.286 |
| **Impacted teeth** | 0.959 | 0.876 | **Root resorption** | 0.167 | 0.000 |
| **Partial edentulism** | 0.828 | 0.849 | **Cleft palate** | 0.000 | 0.167 |
| **Apical periodontitis** | 0.728 | 0.815 | **Radicular cyst** | 0.600 | 0.400 |
| **Root canal treatment** | 0.725 | 0.783 | **Alveolar cleft** | 0.400 | 0.600 |
| **Ethmoid sinusitis** | 0.561 | 0.439 | **Alveolar bone fracture** | 0.600 | 0.400 |
| **Maxillary sinus cyst** | 0.423 | 0.558 | **Inferior turbinate hypertrophy** | 0.250 | 0.000 |
| **Rhinosinusitis** | 0.653 | 0.857 | **Osteomyelitis of the jaw** | 0.250 | 0.000 |
| **Alveolar bone resorption** | 0.660 | 0.638 | **Sphenoid sinus cyst** | 0.333 | 0.333 |
| **Sphenoid sinusitis** | 0.595 | 0.286 | **Paranasal sinus cyst** | 0.500 | 0.500 |
| **TMJ osteoarthritis** | 0.275 | 0.300 | **Dentigerous cyst** | 0.500 | 1.000 |
| **Residual root** | 0.371 | 0.429 | **Deviated nasal septum** | 0.000 | 0.000 |
| **Crown restoration** | 0.548 | 0.548 | **Sialolithiasis** | 1.000 | 1.000 |
| **Orthognathic surgery** | 0.963 | 0.963 | **Tooth dislocation** | 0.000 | 0.000 |
| **Supernumerary tooth/teeth** | 0.565 | 0.609 | **Peri-implant bone loss** | 1.000 | 1.000 |
| **Dental implant placement** | 0.905 | 0.952 | **Fibrous dysplasia** | 0.000 | 1.000 |
| **Tooth structure loss** | 0.550 | 0.550 | **Tooth fusion** | 0.000 | 0.000 |
| **Bone grafting** | 0.375 | 0.438 | **Maxillofacial fracture** | 0.000 | 0.000 |
| **Orthodontic treatment** | 0.333 | 0.583 | **Benign jaw tumor** | 0.000 | 0.000 |
| **Retained primary teeth** | 0.400 | 0.400 | **Osteoma** | 0.000 | 0.000 |

**Table S3. Detailed Subspecialist Preference Ranking of Generated and Human Reports**

| Group | Rank 1 | Rank 2 | Rank 3 | Rank 4 |
|---|---|---|---|---|
| **Radiologists** | | | | |
| AI | 24 (0.080) | 40 (0.133) | 125 (0.417) | 111 (0.370) |
| Novice | 4 (0.013) | 63 (0.210) | 78 (0.260) | 155 (0.517) |
| Intermediate | 58 (0.193) | 135 (0.450) | 82 (0.273) | 25 (0.083) |
| Senior | 214 (0.713) | 62 (0.207) | 14 (0.047) | 10 (0.033) |
| **Clinicians** | | | | |
| AI | 24 (0.240) | 23 (0.230) | 22 (0.220) | 31 (0.310) |
| Novice | 4 (0.040) | 23 (0.230) | 37 (0.370) | 36 (0.360) |
| Intermediate | 17 (0.170) | 23 (0.230) | 28 (0.280) | 32 (0.320) |
| Senior | 55 (0.550) | 31 (0.310) | 13 (0.130) | 1 (0.010) |

Values are presented as count (proportion) of preference rankings assigned by subspecialists. Rankings range from Rank 1 (best) to Rank 4 (worst). Proportions are calculated within each rater cohort (Radiologists: n = 300; Clinicians: n = 100) and sum to 1.00 across ranks for each group (AI, Novice, Intermediate, Senior).

**Table S4. Omission Count and Clinical Significance of Findings and Impressions on AI and Manual Reports Across Four Clinical Subspecialties**

| Subspecialty | Outcome | AI | Novice | Intermediate | Senior | p_value |
|---|---|---|---|---|---|---|
| **Findings** | | | | | | |
| **Ortho** | Count | 1.24[a] | 2.52[a] | 1.52 | 0.71 | 0.000 |
| **Ortho** | Clinical Significance | 0.16[a] | 0.68[a] | 0.4 | 0.21 | 0.000 |
| **OMFS** | Count | 1.70[c] | 2.68 | 1 | 0.64[c] | 0.000 |
| **OMFS** | Clinical Significance | 0.22[a] | 0.73[a] | 0.3 | 0.14 | 0.000 |
| **Perio** | Count | 1.74[c] | 2.43 | 1.13 | 0.50[c] | 0.000 |
| **Perio** | Clinical Significance | 0.26[a] | 0.83[a] | 0.3 | 0.09 | 0.000 |
| **Endo** | Count | 1.40[c] | 2.05 | 1.4 | 0.70[c] | 0.001 |
| **Endo** | Clinical Significance | 0.25 | 0.6 | 0.45 | 0.1 | 0.006 |
| **Impression** | | | | | | |
| **Ortho** | Count | 1.28 | 1.8 | 1.84 | 0.71 | 0.005 |
| **Ortho** | Clinical Significance | 0.36 | 0.56 | 0.5 | 0.08 | 0.003 |
| **OMFS** | Count | 1.65[c] | 2.24 | 1.17 | 0.59[c] | 0.000 |
| **OMFS** | Clinical Significance | 0.52[c] | 0.71 | 0.43 | 0.09[c] | 0.001 |
| **Perio** | Count | 1.78[c] | 2.41 | 1.77 | 0.50[c] | 0.000 |
| **Perio** | Clinical Significance | 0.48[c] | 0.77 | 0.59 | 0.09[c] | 0.000 |
| **Endo** | Count | 1.55[c] | 2 | 1.4 | 0.70[c] | 0.002 |
| **Endo** | Clinical Significance | 0.40[c] | 0.55 | 0.4 | 0.05[c] | 0.008 |

Count denotes the mean number of omission errors per case, and Clinical Significance denotes the mean binary clinical-significance score (0/1), interpretable as a proportion. P_overall was obtained using a Kruskal–Wallis test across the AI, Novice, Intermediate, and Senior groups. Superscripts denote Holm-adjusted pairwise comparisons versus AI based on the Mann–Whitney U test: a, AI versus Novice; b, AI versus Intermediate; c, AI versus Senior. A superscript is shown for both groups only if the corresponding adjusted $P < 0.05$. Abbreviations: OMFS, oral and maxillofacial surgery; Ortho, orthodontics; Perio, periodontology; Endo, endodontics.

**Table S5. Incorrection Count and Clinical Significance of Findings and Impressions on AI and Manual Reports Across Four Clinical Subspecialties**

| Subspecialty | Outcome | AI | Novice | Intermediate | Senior | p_value |
|---|---|---|---|---|---|---|
| **Findings** | | | | | | |
| **Ortho** | Count | 1.44[b,c] | 1.36 | 0.62[b] | 0.62[c] | 0.000 |
| **Ortho** | Clinical Significance | 0.28[a] | 0.64[a] | 0.21 | 0.21 | 0.003 |
| **OMFS** | Count | 1.57[b,c] | 1.45 | 0.30[b] | 0.32[c] | 0.000 |
| **OMFS** | Clinical Significance | 0.57[bc] | 0.36 | 0.17[b] | 0.14[c] | 0.007 |
| **Perio** | Count | 2.04[abc] | 1.35a | 0.48[b] | 0.45[c] | 0.000 |
| **Perio** | Clinical Significance | 0.78[abc] | 0.48a | 0.22[b] | 0.18[c] | 0.000 |
| **Endo** | Count | 1.45[bc] | 1.00 S | 0.40[b] | 0.40[c] | 0.000 |
| **Endo** | Clinical Significance | 0.55[bc] | 0.50 | 0.15[b] | 0.05[c] | 0.001 |
| **Impression** | | | | | | |
| **Ortho** | Count | 1.24[bc] | 1.28 | 0.44[b] | 0.38[c] | 0.000 |
| **Ortho** | Clinical Significance | 0.24[a] | 0.80[a] | 0.16 | 0.12 | 0.000 |
| **OMFS** | Count | 1.13[bc] | 1.09 | 0.43[b] | 0.32[c] | 0.000 |
| **OMFS** | Clinical Significance | 0.48[c] | 0.68 | 0.30 | 0.14[c] | 0.002 |
| **Perio** | Count | 1.48[bc] | 1.04 | 0.50[b] | 0.50[c] | 0.000 |
| **Perio** | Clinical Significance | 0.70[bc] | 0.61 | 0.32[b] | 0.14[c] | 0.000 |
| **Endo** | Count | 1.05[bc] | 0.70 | 0.40[b] | 0.45[c] | 0.025 |
| **Endo** | Clinical Significance | 0.45[bc] | 0.40 | 0.05[b] | 0.10[c] | 0.004 |

Count denotes the mean number of incorrect errors per case, and Clinical Significance denotes the mean binary clinical-significance score (0/1), interpretable as a proportion. P_overall was obtained using a Kruskal–Wallis test across the AI, Novice, Intermediate, and Senior groups. Superscripts denote Holm-adjusted pairwise comparisons versus AI based on the Mann–Whitney U test: a, AI versus Novice; b, AI versus Intermediate; c, AI versus Senior. A superscript is shown for both groups only if the corresponding adjusted P < 0.05. Abbreviations: OMFS, oral and maxillofacial surgery; Ortho, orthodontics; Perio, periodontology; Endo, endodontics.

**Table S6. Detailed Subspecialist Preference Ranking of Collaboration and Manual Reports**

| Group | Rank 1 | Rank 2 | Rank 3 | Rank 4 | Rank 5 | Rank 6 |
|---|---|---|---|---|---|---|
| **Radiologists** | | | | | | |
| Novice | 0 (0.00) | 11 (0.04) | 23 (0.08) | 48 (0.16) | 52 (0.17) | 166 (0.55) |
| Intermediate | 37 (0.12) | 34 (0.11) | 50 (0.17) | 81 (0.27) | 66 (0.22) | 32 (0.11) |
| Senior | 114 (0.38) | 83 (0.28) | 45 (0.15) | 40 (0.13) | 14 (0.05) | 4 (0.01) |
| Co-Novice | 6 (0.02) | 27 (0.09) | 54 (0.18) | 48 (0.16) | 117 (0.39) | 48 (0.16) |
| Co-Intermediate | 41 (0.14) | 58 (0.19) | 86 (0.29) | 48 (0.16) | 49 (0.16) | 18 (0.06) |
| Co-Senior | 107 (0.36) | 87 (0.29) | 43 (0.14) | 31 (0.10) | 21 (0.07) | 11 (0.04) |
| **Clinicians** | | | | | | |
| Novice | 0 (0.00) | 1 (0.01) | 12 (0.12) | 20 (0.20) | 15 (0.15) | 52 (0.52) |
| Intermediate | 4 (0.04) | 12 (0.12) | 15 (0.15) | 14 (0.14) | 26 (0.26) | 29 (0.29) |
| Senior | 27 (0.27) | 33 (0.33) | 9 (0.09) | 17 (0.17) | 13 (0.13) | 1 (0.01) |
| Co-Novice | 15 (0.15) | 8 (0.08) | 19 (0.19) | 22 (0.22) | 19 (0.19) | 17 (0.17) |
| Co-Intermediate | 8 (0.08) | 30 (0.30) | 33 (0.33) | 17 (0.17) | 6 (0.06) | 6 (0.06) |
| Co-Senior | 38 (0.38) | 11 (0.11) | 14 (0.14) | 14 (0.14) | 17 (0.17) | 6 (0.06) |

Values are presented as count (proportion) of preference rankings assigned by subspecialists. Rankings range from Rank 1 (best) to Rank 6 (worst). Proportions are calculated within each rater cohort (Radiologists: $n$ = 300; Clinicians: $n$ = 100) and sum to 1.00 across ranks for each group (Novice, Intermediate, Senior, Co-Novice, Co-Intermediate, Co-Senior).

**Table S7. Omission Count and Clinical Significance of Findings and Impressions on Collaboration and Manual Reports Across Four Clinical Subspecialties**

| Subspecialty | Outcome | Novice | Co-Novice | Intermediate | Co-Intermediate | Senior | Co-Senior | p_value |
|---|---|---|---|---|---|---|---|---|
| **Findings** | | | | | | | | |
| **Ortho** | Count | 2.52 | 1.72 | 1.52 | 0.84 | 0.71 | 0.60 | 0.00 |
| **Ortho** | Clinical Significance | 0.68 | 0.68 | 0.40 | 0.32 | 0.21 | 0.12 | 0.00 |
| **OMFS** | Count | 2.68 | 2.04 | 1.00 | 0.68 | 0.64 | 0.57 | 0.00 |
| **OMFS** | Clinical Significance | 0.73 | 0.61 | 0.30 | 0.23 | 0.14 | 0.04 | 0.00 |
| **Perio** | Count | 2.43 | 2.30 | 1.13 | 1.00 | 0.50 | 0.57 | 0.00 |
| **Perio** | Clinical Significance | 0.83 | 0.78 | 0.30 | 0.43 | 0.09 | 0.00 | 0.00 |
| **Endo** | Count | 2.05 | 2.05 | 1.40 | 1.11 | 0.70 | 0.50 | 0.00 |
| **Endo** | Clinical Significance | 0.60 | 0.60 | 0.45 | 0.42 | 0.10 | 0.05 | 0.00 |
| **Impressions** | | | | | | | | |
| **Ortho** | Count | 1.80 | 1.84 | 1.84[b] | 0.92[b] | 0.71 | 0.92 | 0.00 |
| **Ortho** | Clinical Significance | 0.56 | 0.68 | 0.50 | 0.40 | 0.08 | 0.24 | 0.00 |
| **OMFS** | Count | 2.24 | 1.74 | 1.17 | 0.82 | 0.59 | 0.65 | 0.00 |
| **OMFS** | Clinical Significance | 0.71 | 0.57 | 0.43 | 0.27 | 0.09 | 0.17 | 0.00 |
| **Perio** | Count | 2.41 | 2.32 | 1.77 | 1.17 | 0.50 | 0.70 | 0.00 |
| **Perio** | Clinical Significance | 0.77 | 0.82 | 0.59 | 0.48 | 0.09 | 0.00 | 0.00 |
| **Endo** | Count | 2.00 | 1.95 | 1.40 | 1.11 | 0.70 | 0.50 | 0.00 |
| **Endo** | Clinical Significance | 0.55 | 0.55 | 0.40 | 0.37 | 0.05 | 0.10 | 0.00 |

Values are reported as means. *Count* denotes the mean number of omission errors per case; *Clinical Significance* denotes the proportion of omissions labeled clinically significant. *P_value* indicates the overall between-group difference within each subspecialty (Novice, Co-Novice, Intermediate, Co-Intermediate, Senior, Co-Senior) assessed using a two-sided Kruskal–Wallis test. Abbreviations: OMFS, oral and maxillofacial surgery; Ortho, orthodontics; Perio, periodontology; Endo, endodontics.

**Table S8. Incorrection Count and Clinical Significance of Findings and Impressions on Collaboration and Manual Reports Across Four Clinical Subspecialties**

| Subspecialty | Outcome | Novice | Co-Novice | Intermediate | Co-Intermediate | Senior | Co-Senior | p_value |
|---|---|---|---|---|---|---|---|---|
| **Findings** | | | | | | | | |
| **Ortho** | Count | 1.36 | 1.52 | 0.62 | 0.48 | 0.62 | 0.60 | 0.000 |
| **Ortho** | Clinical Significance | 0.64 | 0.56 | 0.21 | 0.20 | 0.21 | 0.20 | 0.000 |
| **OMFS** | Count | 1.45 | 1.43 | 0.30 | 0.50 | 0.32 | 0.39 | 0.000 |
| **OMFS** | Clinical Significance | 0.36 | 0.43 | 0.17 | 0.18 | 0.14 | 0.13 | 0.065 |
| **Perio** | Count | 1.35 | 1.61 | 0.48 | 0.52 | 0.45 | 0.48 | 0.000 |
| **Perio** | Clinical Significance | 0.48 | 0.39 | 0.22 | 0.22 | 0.18 | 0.09 | 0.031 |
| **Endo** | Count | 1.00 | 1.20 | 0.40 | 0.84 | 0.40 | 0.35 | 0.005 |
| **Endo** | Clinical Significance | 0.50 | 0.50 | 0.15 | 0.26 | 0.05 | 0.15 | 0.002 |
| **Impressions** | | | | | | | | |
| **Ortho** | Count | 1.28 | 0.96 | 0.44 | 0.40 | 0.38 | 0.56 | 0.000 |
| **Ortho** | Clinical Significance | 0.80 | 0.48 | 0.16 | 0.16 | 0.12 | 0.16 | 0.000 |
| **OMFS** | Count | 1.09 | 1.17 | 0.43 | 0.50 | 0.32 | 0.39 | 0.000 |
| **OMFS** | Clinical Significance | 0.68 | 0.43 | 0.30 | 0.18 | 0.14 | 0.13 | 0.000 |
| **Perio** | Count | 1.04 | 0.86 | 0.50 | 0.61 | 0.50 | 0.52 | 0.114 |
| **Perio** | Clinical Significance | 0.61 | 0.27 | 0.32 | 0.26 | 0.14 | 0.04 | 0.001 |
| **Endo** | Count | 0.70 | 0.95 | 0.40 | 0.95 | 0.45 | 0.40 | 0.066 |
| **Endo** | Clinical Significance | 0.40 | 0.45 | 0.05 | 0.42 | 0.10 | 0.15 | 0.005 |

Values are reported as means. *Count* denotes the mean number of omission errors per case; *Clinical Significance (CS)* denotes the proportion of omissions labeled clinically significant (mean of binary 0/1 annotations). *p_value* indicates the overall between-group difference within each subspecialty (Novice, Co-Novice, Intermediate, Co-Intermediate, Senior, Co-Senior) assessed using a two-sided Kruskal–Wallis test. Abbreviations: OMFS, oral and maxillofacial surgery; Ortho, orthodontics; Perio, periodontology; Endo, endodontics; CS, clinical significance.

**Table S9. Diseases and treatment status in CBCT**

| Impression | Radiographic features | ICD-10 |
|---|---|---|
| **Apical periodontitis** | Radiolucency at the apex of a nonvital tooth, initially presenting as widening of the apical periodontal ligament space, followed by gradual loss of the apical lamina dura. | K04.5 |
| **Impacted Tooth** | Teeth that fail to erupt into their normal position during the normal eruption period are referred to as impacted teeth. Impacted mandibular and maxillary third molars are the most common. | K01.1 |
| **Malocclusion** | Abnormalities in tooth alignment and occlusion；abnormalities in jaw size and position. | K07 |
| **Supernumerary tooth** | Fully developed unerupted supernumerary teeth are surrounded by a dental follicle, while immature unerupted ones appear as tooth buds at various stages of development.Erupted supernumerary teeth may either align with the dental arch or be positioned labially or lingually relative to adjacent teeth. | K00.1 |
| **Partial edentulism** | The absence of one to several teeth in the dental arch, while some natural teeth still remain in the oral cavity. | K08.1 |
| **Post–root canal treatment** | Radiodense material is seen within the root canal space; no significant radiolucency is observed in the periapical region. | Z92.8 |
| **Altered TMJ joint space** | Widening or narrowing of the anterior and posterior superior joint spaces of the temporomandibular joint; widening or narrowing of the entire joint space. These changes in joint space may occur symmetrically in both joints or asymmetrically. | K07.6 |
| **Dental Caries** | Caries can affect any tooth surface, appearing as a radiolucent area whose size varies with severity. | K02 |
| **Maxillary sinusitis** | Maxillary sinus mucosal thickening or soft tissue opacification accompanied by bony wall thickening and sclerosis, with the affected sinus cavity either normal or reduced in volume. | J32.0 |
| **Tooth fracture** | A tooth fracture line appears as an irregular, thin, linear radiolucent image, with possible slight displacement between the fractured ends. Fractures can be classified as crown fractures, root fractures, or combined crown-root fractures. | S02.5 |
| **Post-orthognathic surgery** | After maxillary and mandibular osteotomy and repositioning, the retention titanium plates and screws show no signs of displacement or loosening. Multiple micro-implants are placed in the upper and lower alveolar processes, and fixed orthodontic appliances are attached to the upper and lower dental arches. | Z92.8 |
| **Post-implant surgery** | Tooth loss with a dental implant placed in the corresponding area. A mass-like/lumpy slightly hyperdense implant material is visible on the labial/buccal side, retained by titanium screws. | Z92.8 |
| **Alveolar bone resorption** | Bone loss around teeth, which may involve a single tooth or a few teeth (localized) or affect all teeth (generalized), with various patterns of bone resorption. | K05.3 |
| **Maxillary sinus cyst** | Dome-shaped soft tissue mass in the maxillary sinus, lacking cortication and associated sinus wall changes; most commonly located on the sinus floor or wall, and may be multiple. | J34.1 |
| **Partial maxillofacial bone resection surgery** | Irregular bone defect is seen within the jawbone, with well-defined margins. The buccal/lingual/palatal/labial cortical bone is partially discontinuous. | Z92.8 |
| **TMJ osteoarthritis** | Condylar size reduction, often involving the superior surface, may be accompanied by sclerosis, flattening, erosions, osteophytes, subchondral cysts, joint space narrowing, and disc displacement. | K07.6 |
| **Post-orthodontic treatment** | Fixed orthodontic appliances are attached to the upper and lower teeth. | Z92.8 |
| **Paranasal sinusitis** | Paranasal sinus mucosal thickening or soft tissue opacification accompanied by bony wall thickening and sclerosis, with the affected sinus cavity either normal or reduced in volume. | J32.4 |
| **Residual root** | The clinical crown is missing, with the root remaining in the alveolar socket; the root may be intact in length and morphology or show partial resorption. | K08.3 |
| **Tooth structure defect** | Due to caries, fractures, wear, or other reasons, the hard tissues of the tooth exhibit X-ray low-density radiolucency or morphological disruption, which may be accompanied by pulp chamber exposure and periapical inflammatory changes. | K03.9 |
| **Post-bone grafting** | Slightly hyperdense implant material is seen. | Z92.8 |
| **Root resorption** | Irregular loss of tooth structure on the internal (pulpal) or external (periodontal) surface of portions of the tooth located beneath bone or soft tissue; the extent varies with lesion progression. | K03.3 |
| **Radicular cyst** | Unilocular, round, well-corticated radiolucency located at the root apex. | K04.8 |
| **Crown restoration** | A radiopaque restoration is seen in the crown portion. | Z92.8 |
| **Maxillofacial fracture** | Low-density fracture lines, abnormal radiodense lines, trabecular distortion and buckling, free bone fragments, compression/deformation of the facial bones, and suture separation are observed in the maxillofacial bones. | S02 |
| **Retained primary tooth** | Retained primary tooth, with or without evidence of the permanent successor. | K00.6 |
| **Ethmoid sinusitis** | Ethmoid sinus mucosal thickening or soft tissue opacification accompanied by bony wall thickening and sclerosis, with the affected sinus cavity either normal or reduced in volume. | J32.2 |
| **Post-extraction** | Low-density radiolucency at the original root site, with well-defined borders and no evidence of root structure. | Z92.8 |
| **Fixed bridge restoration** | Restored with a crown and bridge from tooth # to tooth #. | Z92.8 |
| **Dentigerous cyst** | A unilocular radiolucency associated with the crown of an unerupted tooth, typically well-defined with a sclerotic border; however, infected cysts may exhibit ill-defined margins. | K09.0 |
| **Paranasal sinus cyst** | Dome-shaped soft tissue masses in the paranasal sinuses, lacking cortication and associated sinus wall changes; may be multiple. | J34.1 |
| **Inferior turbinate hypertrophy** | Hypertrophy of the inferior nasal concha: increased overall volume due to mucosal thickening, bony hyperplasia, or both. | J34.3 |
| **Alveolar ridge cleft** | Clefts of the alveolus most commonly occur between the lateral incisor and canine. They are frequently associated with missing or supernumerary teeth. The defect often extends into the nasal cavity, resulting in asymmetry of the nasal floor and nasal cavity contents. | K08.8 |
| **Post-fracture surgery** | Internal fixation devices are in good position without loosening or breakage. The fracture fragments show satisfactory alignment and apposition, with the fracture line gradually becoming blurred and disappearing. | Z92.8 |
| **Dentoalveolar Fractures** | Fracture lines are often transverse, oblique, or longitudinal, appearing as irregular, uneven linear radiolucencies. Frequently accompanied by dental injuries. | S02.8 |
| **Jaw benign tumor** | Jaw tumors and tumor-like lesions can be classified as odontogenic or non-odontogenic. Among jaw tumors, the majority are benign, with benign odontogenic tumors being the most prevalent. | D16.4<br>D16.5 |
| **Sphenoid sinusitis** | Sphenoid sinus mucosal thickening or soft tissue opacification accompanied by bony wall thickening and sclerosis, with the affected sinus cavity either normal or reduced in volume. | J32.3 |
| **Sialolithiasis** | Sialoliths typically present as single or multiple ovoid radiopaque masses within the salivary gland duct. The | K11.5 |
| | submandibular gland is the most common site of occurrence. | |
| **Craniofacial cystic lesions** | Round or oval radiolucent lesions within the jawbone. Cystic lesions of the jaws are classified as odontogenic or non-odontogenic cysts. The former mainly include inflammatory periapical (radicular) cysts, odontogenic keratocysts, dentigerous cysts, etc.; the latter are mostly developmental cysts, such as the nasopalatine duct cyst. | K09 |
| **Osteomyelitis of the jaw** | Osteomyelitis of the jaws may present as lytic, sclerotic, or mixed patterns, accompanied by sequestra and laminated periosteal new bone formation. | K10.2 |
| **Fused tooth** | An enlarged crown, with or without clefting; a single large or partially divided pulp chamber; and a single large root or separate roots. | K00.2 |
| **Sphenoid sinus cyst** | Dome-shaped soft tissue masses in the sphenoid sinus, lacking cortication and associated sinus wall changes; may be multiple. | J34.1 |
| **Change in the physiological curvature of the cervical spine** | Loss of the normal forward convex curvature of the cervical spine, presenting as straightening, kyphosis, or an "S"-shaped double curve. | M48.9 |
| **Osteoma** | Radiographically, osteomas appear as circumscribed sclerotic masses. | D16 |
| **Cleft palate** | Discontinuity of the hard palate at the midline, off-midline, or bilaterally, presenting as a radiolucent band with well-defined or sclerotic margins. | Q35.1 |
| **Luxation** | Extrusive luxation: Crown displaced lingually and coronally, root displaced facially.<br>Intrusive luxation: Tooth displaced apically, with apex positioned higher than that of the contralateral tooth; partial or total obliteration of the periapical PDL space.<br>Lateral luxation: Displacement direction varies depending on the traumatic force. | S03.2 |
| **Jaw cyst** | Round or oval radiolucent lesions within the jawbone. Cystic lesions of the jaws are classified as odontogenic or non-odontogenic cysts. The former mainly include inflammatory periapical (radicular) cysts, odontogenic keratocysts, dentigerous cysts, etc.; the latter are mostly developmental cysts. | K09 |
| **Deviated Nasal Septum** | Marked deviation from midline position of nasal septum, creating nasal fossae of unequal size | J34.2 |
| **Peri-implant bone loss** | Peri-implant alveolar bone resorption, blurring and loss of the lamina dura at the alveolar crest, and reduction in alveolar ridge height. | K05.3 |
| **Dentinogenesis imperfecta** | The teeth exhibit bulbous crowns, cervical constrictions, thin roots, and obliteration of pulp chambers and root canals. | K00.5 |
| **Fibrous dysplasia** | The primary radiographic feature is a fine, ground-glass opacification, and the lesions of fibrous dysplasia are typically not well demarcated. | K10.8 |
| **Ossifying Fibroma** | Round or oval, well-defined, expansile mass with a corticated border and variable internal radiopacity; the internal aspect may be granular. | D16 |
| **Jaw fracture** | Radiolucent, noncorticated lines in the jaw, with variable diastasis, angulation, and comminution. | S02.4<br>S02.6 |
| **Frontal sinus** | Frontal sinus mucosal thickening or soft tissue opacification accompanied by bony wall thickening and sclerosis, with the affected sinus cavity either normal or reduced in volume. | J32.1 |

Table S10. Detailed Description of Quality Evaluation Metrics

| Aspect | Score | Description |
|---|---|---|
| **Factual consistency** | | |
| **Best** | 1 | The report is completely accurate. |
| | 2 | The report provides accurate diagnosis of critical diseases with some errors in noncritical ones. |
| | 3 | The report contains misdiagnoses or over-diagnoses of critical diseases. |
| | 4 | The report contains completely incorrect diagnoses. |
| **Coherence** | | |
| **Best** | 1 | The report has the correct format, clear logic, and concise phrasing. |
| | 2 | The report has the correct format but with redundant or oversimplified expressions. |
| | 3 | The report has an incorrect format and logical errors. |
| | 4 | The report is completely unreadable. |
| **Medical safety** | | |
| **Best** | 1 | The report can be signed off directly and is considered safe for patient care. |
| | 2 | The report contains minor flaws that can be corrected with minimal edits before signing. |
| | 3 | The report contains significant errors requiring substantial revisions. |
| | 4 | The report is an entirely unusable report that poses a risk to patient safety and requires extensive rework or a complete rewrite. |
| **Clinical use** | | |
| **Best** | 1 | The report fully addresses the clinical question. It provides a definitive diagnosis of the target lesion, accurately describes lesion extent and margins, and the relationship to adjacent anatomical structures. Incidental non-target findings are accurately identified and clearly reported. |
| | 2 | The report partially addresses the clinical question. It provides a definitive diagnosis of the target lesion but descriptions of lesion extent and margins, and the relationship to adjacent anatomical structures are incomplete. Non-target findings are reported, but some descriptions are vague or potentially ambiguous. |
| | 3 | The report minimally addresses the clinical question and contains unclear descriptions. The diagnosis and characterization of lesion are ambiguous. Key anatomical relationships are inadequately described. |
| | 4 | The report barely addresses the clinical question and contains inaccurate descriptions. The diagnosis and characterization of lesion are incorrect or contradictory, and critical anatomical relationships are misidentified or omitted. |